# Adaptive Covariance and Quaternion-Focused Hybrid Error-State EKF/UKF for Visual-Inertial Odometry

Ufuk Asil[1] · Efendi Nasibov[1]

## Abstract

This study presents an innovative hybrid Visual-Inertial Odometry (VIO) method for Unmanned Aerial Vehicles (UAVs) that is resilient to environmental challenges and capable of dynamically assessing sensor reliability. Built upon a loosely coupled sensor fusion architecture, the system utilizes a novel hybrid Quaternion-focused Error-State EKF/UKF (Qf-ES-EKF/UKF) architecture to process inertial measurement unit (IMU) data. This architecture first propagates the entire state using an Error-State Extended Kalman Filter (ESKF) and then applies a targeted Scaled Unscented Kalman Filter (SUKF) step to refine only the orientation. This sequential process blends the accuracy of SUKF in quaternion estimation with the overall computational efficiency of ESKF. The reliability of visual measurements is assessed via a dynamic sensor confidence score based on metrics, such as image entropy, intensity variation, motion blur, and inference quality, adapting the measurement noise covariance to ensure stable pose estimation even under challenging conditions. Comprehensive experimental analyses on the EuRoC MAV dataset demonstrate key advantages: an average improvement of 49% in position accuracy in challenging scenarios, an average of 57% in rotation accuracy over ESKF-based methods, and SUKF-comparable accuracy achieved with approximately 48% lower computational cost than a full SUKF implementation. These findings demonstrate that the presented approach strikes an effective balance between computational efficiency and estimation accuracy, and significantly enhances UAV pose estimation performance in complex environments with varying sensor reliability.

## 1 Introduction

VIO is a critical technology for position estimation and navigation of autonomous platforms, especially in indoor environments where Global Navigation Satellite Systems (GNSS) are unavailable or unreliable. In systems with high dynamics, such as Unmanned Aerial Vehicles (UAVs), which are increasingly employed for complex tasks like autonomous navigation and path planning [1], the performance of VIO can be adversely affected by various environmental factors. Challenges, such as motion blur, sudden illumination changes, and insufficient visual texture, push the accuracy and stability limits of traditional odometry methods [2–5].

The existing literature offers innovative solutions in both visual and inertial data processing pipelines to overcome these challenges. In the field of visual processing, deep learning-based architectures running on GPUs exhibit more robust performance in challenging conditions. This includes not only feature matching but also other perception tasks like object detection on UAVs [6]. However, these methods often increase computational complexity, creating a critical latency-accuracy trade-off in real-time applications, which researchers actively try to mitigate [7–11].

In the inertial measurement channel, the Extended Kalman Filter (EKF), widely used due to its compatibility with low-cost Micro-Electro-Mechanical Systems (MEMS)-based sensors, tends to accumulate modeling errors in

 

high-dynamic and non-linear motion profiles. In this context, iterative optimization techniques such as Iterative-EKF [12] and sigma-point-based filtering paradigms like the Unscented Kalman Filter (UKF) [13], offered as alternatives to EKF, promise performance improvements, particularly in the calibration and state estimation stages of visual–inertial sensor fusion. Increased processing capacity due to recent hardware advancements highlights the capability of UKF and its variants to model non-linear system dynamics more effectively, making it a promising research area for VIO systems.

Tightly coupled sensor fusion systems in the literature theoretically offer high-accuracy potential. However, in practice, they have disadvantages, such as high computational complexity and increased resource consumption. Furthermore, critical errors or measurement anomalies originating from one sensor in such systems can adversely affect the performance of the entire system due to the nature of model-based fusion architecture. In contrast, the success level achieved by graph-based visual odometry methods in recent times demonstrates that results close to multi-sensor fusion can be obtained even with camera sensors alone. These developments bring the advantages of loosely coupled fusion architectures, such as modularity and computational efficiency, back to the forefront. Loosely coupled systems allow for parallel processing of heterogeneous sensors, thereby preserving computational efficiency and offering the opportunity to leverage the advantages of each sensor in a situation-specific manner. Especially, in dynamic environmental conditions, managing information obtained from different sensor channels through weighted integration or situational switching mechanisms can reduce the vulnerabilities of tightly coupled systems and provide a more modular and adaptive navigation framework. In this paradigm, hybrid fusion approaches are attracting attention as an innovative research area [14–26].

In recent years, a notable increase has been observed in studies addressing the integration of heterogeneous sensors beyond IMUs and cameras. This includes technologies like ultra-wideband (UWB) radar, Light Detection and Ranging (LiDAR), and next-generation communication systems, such as 5 G and 6 G, which can be leveraged for both communication and positioning [27–30]. However, these sensors, much like visual sensors, also have vulnerabilities dependent on environmental conditions. For example, LiDAR sensors can produce erroneous point clouds under atmospheric disturbances (fog, rain) due to photon scattering and reflection losses, while 5 G-based positioning systems can be adversely affected by signal delays caused by multipath propagation and physical obstacles. In the context of heterogeneous sensor fusion, a systematic analysis of the relative advantages and limitations of each sensor depending on environmental parameters necessitates the design of a dynamic optimization mechanism based on adaptive weighting principles. Such a mechanism can enhance the overall robustness and fault tolerance of the system by blending multi-sensor outputs in real-time according to reliability metrics.

This need in the literature encourages research focused particularly on switching strategies and situational decision-making architectures. It is important to manage inter-sensor transitions with hybrid methods triggered by online monitoring of environmental parameters. This management holds the potential to balance resource efficiency and performance. Such studies are expected to offer methodological contributions that will strengthen the continuous navigation capabilities of autonomous systems, especially in complex and variable environments.

The adaptive and hybrid VIO approach presented in this study focuses on enhancing the pose estimation performance of UAVs in challenging and dynamic environments. The original contributions of the proposed system to the literature are listed below:

- **Advanced State Estimation with Hybrid Qf-ES-EKF/UKF:** An innovative hybrid Qf-ES-EKF/UKF architecture is presented for processing IMU data. This architecture enhances orientation estimation through a two-stage process: first, the entire state is propagated using an efficient ESKF; then, the orientation component is specifically refined using the more accurate SUKF. This approach provides precise localization on the manifold while maintaining computational efficiency. This approach, in loosely coupled integration with graph-based visual odometry, provides more precise and accurate localization, especially in rotation estimations and on the manifold.
- **Dynamic and Adaptive Sensor Fusion with CASEF:** The reliability of visual measurements is dynamically assessed using various quality metrics (e.g., image entropy, intensity variation, motion blur, and visual optimization error). Its impact on the measurement noise covariance is adaptively modulated by the original Clipped Adaptive Saturation Exponential Function (CASEF). CASEF enables precise control of the saturation characteristic via an adjustable $s$ parameter and offers the ability to mimic the behavior of various conventional activation functions. This reduces the impact of low-quality visual data, enabling situational switching between sensors.

- **Proven Performance in High-Dynamic Environments:** The proposed system's robustness and accuracy were validated on the EuRoC MAV dataset, demonstrating significant advantages. It improves position accuracy by an average of 49% in challenging scenarios and rotation accuracy by 57% over ESKF, all while delivering SUKF-level performance at approximately 48% lower computational cost.

The remainder of the paper is organized as follows: Sect. 2 presents a review of the related literature. Section 3 details the proposed method. Section 4 shares the experimental results and evaluations. Section 5 summarizes the results and key findings. Future work is suggested in Sect. 6.

## 2 Related Work

In recent years, VIO has become one of the significant research areas in robotics and computer vision, particularly in the context of autonomous navigation systems, driven by increasing interest. Studies in this field focus on developing real-time and robust algorithms that enable robots and other autonomous systems to estimate their position and orientation with high accuracy and reliability. This section comprehensively reviews the VIO literature, systematically examining filtering and optimization-based methods, visual data processing procedures, next-generation sensor configurations, and landmark studies in method evaluation, as well as current methods similar to the proposed approach. This section provides a comprehensive review of the VIO literature. Filtering and optimization-based methods, visual data processing procedures, and next-generation sensor configurations are examined. Additionally, landmark studies in the evaluation of methods and current approaches similar to the one proposed are systematically reviewed.

### 2.1 Filtering Techniques Based on Inertial Measurements

The EKF estimates by linearizing non-linear systems at each step using a first-order Taylor series. This method is a fundamental approach in fields such as VIO and Simultaneous Localization and Mapping (SLAM). EKF is effective on mobile platforms with low processing power. Due to this effectiveness, it formed the basis for pioneering algorithms such as the tightly coupled MSCKF [31] and the loosely coupled Weiss VIO [32], which were among the first real-time VIO algorithms. Its success in modeling dynamics, especially in mobile systems, has been widely emphasized in the literature [33].

However, the EKF's first-order approximation can be insufficient for the precise estimation of highly non-linear structures, such as rotations represented by quaternions. Such structures often involve complex and trigonometric relationships. To overcome these limitations, alternative approaches like the Second-Order Extended Kalman Filter (EKF2), which uses a second-order Taylor series expansion, have been developed. Although EKF2 potentially offers more accurate estimations than EKF, its use in real-time mobile applications has been very limited due to its excessive computational cost [34].

The Unscented Kalman Filter (UKF) models non-linear transformations using sigma points without requiring a linearization step. This allows it to offer higher accuracy in complex scenarios compared to the EKF. However, comparisons have shown that this accuracy advantage requires three-to-four times more computation time [34, 35]. In response to this computational load problem, various UKF variants have been developed in the literature, aiming both to reduce this load and to increase estimation accuracy.

Among the approaches aimed at minimizing computational load, the Additive Unscented Kalman Filter (AUKF), where process and measurement noises are not augmented to the state vector, stands out. In AUKF, since sigma points are generated only around the state vector, not including the noise dimension in processing significantly reduces computational and memory load [36]. The Central-Difference KF (CDKF), developed for a similar purpose, uses Stirling's central difference interpolation to derive sigma points. This allows it to offer equivalent or better performance than UKF while also reducing computational complexity [37, 38]. Furthermore, the Cubature Kalman Filter (CKF) method uses third-order spherical-radial cubature rules to approximate integrals. This approach produces results with accuracy similar to UKF, eliminates the need for parameter tuning, and improves filter stability [39].

The Scaled UKF (SUKF) was developed by Wan and van der Merwe by adding three scaling parameters ($\alpha$, $\beta$, and $\kappa$) to the Unscented Transformation (SUT). This allows the distribution of sigma points to be better adapted to the input covariance and enhances numerical stability [40]. The Square-Root UKF (SR-UKF) uses the Cholesky factor of the covariance matrix in propagation steps, ensuring that the covariance remains positive definite and thus further increases numerical stability [41]. The Gaussian Hermite Kalman Filter (GHKF) increases theoretical accuracy by calculating the integral of non-linear functions with higher precision using the Gauss Hermite quadrature method; however, its practical use is limited, because its computational cost increases exponentially with the dimension of the state space [42].

One of the more recent UKF variants, the Maximum Correntropy UKF (MC-UKF), is based on the maximum correntropy criterion for situations where measurement and process noises deviate from Gaussian distributions. This makes the filter more robust against outliers [43]. The H∞UKF model, on the other hand, guarantees filter stability by minimizing the worst-case error energy in estimation, even under uncertain or adversarial noise conditions; this approach is particularly effective in applications where sudden changes in sensor noise are expected [44]. However, both methods can produce conservative estimation results and require careful tuning of hyperparameters for optimal performance.

The Ensemble Kalman Filter (EnKF) uses a randomly sampled ensemble instead of deterministic sigma points to more accurately reflect statistical diversity and uncertainties in high-dimensional state spaces. However, the random sampling-based nature of EnKF and its requirement for a large ensemble size make it less preferred compared to UKF variants for low-dimensional VIO problems [45].

For example, while SUKF or SR-UKF is preferred for mobile unmanned aerial vehicles, the use of MC-UKF or H∞UKF is more suitable in environments with high uncertainty, such as industrial robotic arms. Experimental comparisons in the literature show that these choices are directly related to system requirements and environmental conditions.

As an example of the search for hybrid filtering in the literature, the Lie group-based UKF-LG (Unscented Kalman Filter on Lie Groups) method proposed by Brossard et al. [46] can be cited. This work defines the system state within a tightly coupled sensor fusion framework. This state consists of a high-dimensional Lie group element ($SE_{2+p}(3)$) that includes the robot's "pose (position and $SO(3)$ orientation), velocity, and positions of 3D map points," augmented with a vector containing IMU sensor biases. UKF-LG uses sigma points in both propagation and update steps, taking into account this modeled Lie group structure. Thus, the complexity of analytical Jacobian calculations required by EKF-based invariant approaches is avoided, increasing flexibility in system modeling. The authors reported that their proposed Right-UKF-LG variant exhibits competitive performance compared to traditional UKF solutions. This philosophy is important in loosely coupled fusion systems where state vector components are decomposable or have different degrees of dynamic/measurement non-linearity. This is particularly relevant for UAVs, which must execute complex tasks like real-time path planning and multi-agent coordination, often under strict time or resource constraints [47–50]. Such applications demand robust state estimation, making hybrid filters that balance accuracy and efficiency highly desirable. This philosophy combines the precision of higher-order approaches like UKF for specific sub-states (e.g., $SO(3)$ orientation manifolds) while concurrently leveraging the computational efficiency of EKF-derivative methods for other more linearly behaving components, thereby forming a strong foundation for targeted and highly adaptable compact filter designs.

## 2.2 Visual Odometry and Optimization Techniques

In VIO systems, the effectiveness of the visual front-end is directly dependent on the quality of the features used and the consistent matching of these features between consecutive images. Historically, MSCKF [31], one of the first real-time VIO algorithms, pioneered practical mobile applications. It achieved this by combining Shi Tomasi corner detection with the Lucas Kanade (KLT) optical flow algorithm for low computational cost. Subsequently, Weiss VIO [32], one of the first loosely coupled VIO methods, adopted FAST feature detection and KLT matching. This provided sufficient accuracy and speed for on-board applications in micro-UAVs.

The visual-inertial variant of the ORB-SLAM family, VI-ORB-SLAM [51], standardized ORB (Oriented FAST + Rotated BRIEF) features in all tracking, matching, and loop closure processes, successfully minimizing scale drift. The robustness of loop closure, a critical component for long-term drift correction in SLAM, has been further enhanced by recent lightweight deep learning approaches specifically designed for this task [52]. This method processed visual features in an integrated framework, providing high-accuracy matches.

In recent years, the development of deep learning-based methods has led to significant improvements in the visual front-end of VIO systems. In this context, methods, such as SuperPoint [53], D2-Net [54], and R2D2 [55], have produced more stable and consistent results in complex scenes compared to traditional feature extractors. SuperGlue [7], which particularly uses a Transformer architecture, significantly increases visual matching accuracy with its context-based matching strategy.

More recently, ALIKED [56], with its partially differentiable structure, achieved sub-pixel accuracy feature detection while attaining high processing performance suitable for real-time applications. Furthermore, LightGlue [8], a version of SuperGlue with reduced computational load and dynamic adjustability, has become more suitable for mobile and embedded systems due to its ability to automatically regulate computational complexity according to the difficulty level.

In conclusion, ALIKED's high-speed and consistent feature extraction, combined with LightGlue's flexible and adaptive matching strategy, points to the advanced level reached today from the Shi Tomasi corner detection and simple optical flow matching methods used in the traditional MSCKF-based approach. These contemporary methods, while maintaining processing efficiency, exhibit higher matching performance, especially under challenging lighting conditions and appearance changes, thereby increasing the reliability of visual-based odometry systems.

In VIO and SLAM systems, in addition to filter-based approaches, optimization-based methods are also widely used. Algorithmic advancements in this field in recent years have led to significant progress. Now, even optimizations based solely on visual inputs can operate with high accuracy and flexibility. These developments allow optimized visual measurements to be used in filter-based state updates in loosely coupled sensor fusion approaches. This, in turn, enhances the overall performance of visual odometry systems.

Optimization-based techniques are generally classified as Batch, Sliding Window, and incremental methods. Batch Optimization offers the highest accuracy by solving all measurement and state parameters at once, but it is impractical for real-time applications due to the requirement of large matrix inversions and high memory usage. Bundle Adjustment (BA), the fundamental technique in this class, simultaneously optimizes camera poses and 3D points and plays a critical role in scale correction and precise localization in systems like VI-ORB-SLAM [51] and VINS-Mono [57].

Sliding Window Optimization was developed to reduce the high processing load imposed by Batch methods for real-time applications. This method optimizes only recent states within a window, marginalizing old information (removing it from the system of equations and blending it with measurements), thereby reducing the computational load. This technique is frequently used in tightly coupled systems, such as OKVIS [17, 58], VINS-Mono [57], and Kimera [59].

Incremental Smoothing and Mapping (iSAM/iSAM2) methods, on the other hand, are based on the principles of factor graphs and Bayes nets, allowing new data to be added incrementally. These methods offer an ideal solution for real-time applications, especially due to their efficient updates based on Schur complement, which provides linear-time computational performance [60]. The factor graphs underlying these techniques graphically represent the solution of non-linear least squares (NLSQ) problems. Thanks to factor graphs, large-scale optimization problems are solved more efficiently and stably using block diagonal structure and the Schur complement.

To increase robustness against measurement noise, robust cost functions, such as Huber, Tukey, and Geman McClure, are widely used. These functions, especially in systems like VINS-Mono [57] and VI-ORB-SLAM [51], reduce the impact of measurement errors, thereby enhancing overall performance.

Open-source libraries like PySLAM facilitate the practical application and integration of factor graphs and incremental optimization methods, providing researchers with a rapid prototyping and testing environment [61].

The development of optimization techniques has evolved from Batch methods toward Sliding Window and incremental optimization methods. These methods have come to the forefront, especially in real-time applications, by providing a critical balance between computational load and accuracy.

## 2.3 Adaptive Sensor Fusion Approaches

The effectiveness of different sensors in analyzing environmental changes varies according to motion scenarios. Therefore, identifying the conditions under which each sensor provides the most reliable results and prioritizing sensor data accordingly plays a critical role in enhancing the overall estimation performance of the system. For example, a KLT feature tracker, which offers the advantages of low hardware requirements and fast computation, may be inadequate in tracking sudden rotations, whereas gyroscope sensors are more successful in detecting high-speed rotational movements. In the literature, approaches that combine the strengths of different sensors by weighting sensor data with prior information based on the nature of the motion are frequently used as an effective strategy to improve position and orientation estimation accuracy, especially in complex motion scenarios [62–67]. However, visual sensors, in particular, can be adversely affected by factors, such as challenging illumination conditions, occlusion, and motion blur. In tightly coupled sensor fusion architectures, raw data from sensors are processed simultaneously. This situation can make it difficult to dynamically adjust sensor confidence scores when visual distortions or temporary interruptions occur. Furthermore, an error in one sensor can negatively impact others. Various solutions to these visual challenges have been proposed in the literature. Kim et al. [68] presented a visual SLAM approach resilient to illumination changes using a stereo camera system with different exposure levels and restoring over/underexposed regions by masking them through entropy-based analysis and using data from the other camera. Han et al. [69], on the other hand, developed a novel camera control method that estimates motion speed with optical flow, adapts exposure time according to scene dynamics, and enhances image quality even in HDR environments with a special metric combining image gradient/entropy, thereby reducing the adverse effects of motion blur on visual odometry. This study emphasizes an approach to the motion blur problem with mechanical solutions beyond automatic exposure [69]. These and similar solutions in the literature (e.g., entropy analyses, optical flow-assisted exposure control, and dynamic masking) aim to enhance visual odometry performance. However, these solutions often require complex algorithms or additional hardware. In contrast, in VIO systems, IMU sensors stand out as an independent and reliable data source when visual inferences are erroneous or unreliable. The IMU supports the overall system accuracy by making parallel inferences unaffected by visual distortions. This principle of leveraging the best available sensor for a given situation is also seen in specific tasks, such as using visual markers and reinforcement learning for precise autonomous drone landing [70]. In this context, for filter-based loosely coupled sensor fusion approaches like EKF or UKF, dynamically adjusting the covariance values of sensors based on quality or information metrics directly obtained from images (e.g., blur level, feature count/quality) offers a cost-effective and practical solution. This approach is noteworthy for its applicability with standard hardware and its ease of testing on global datasets.

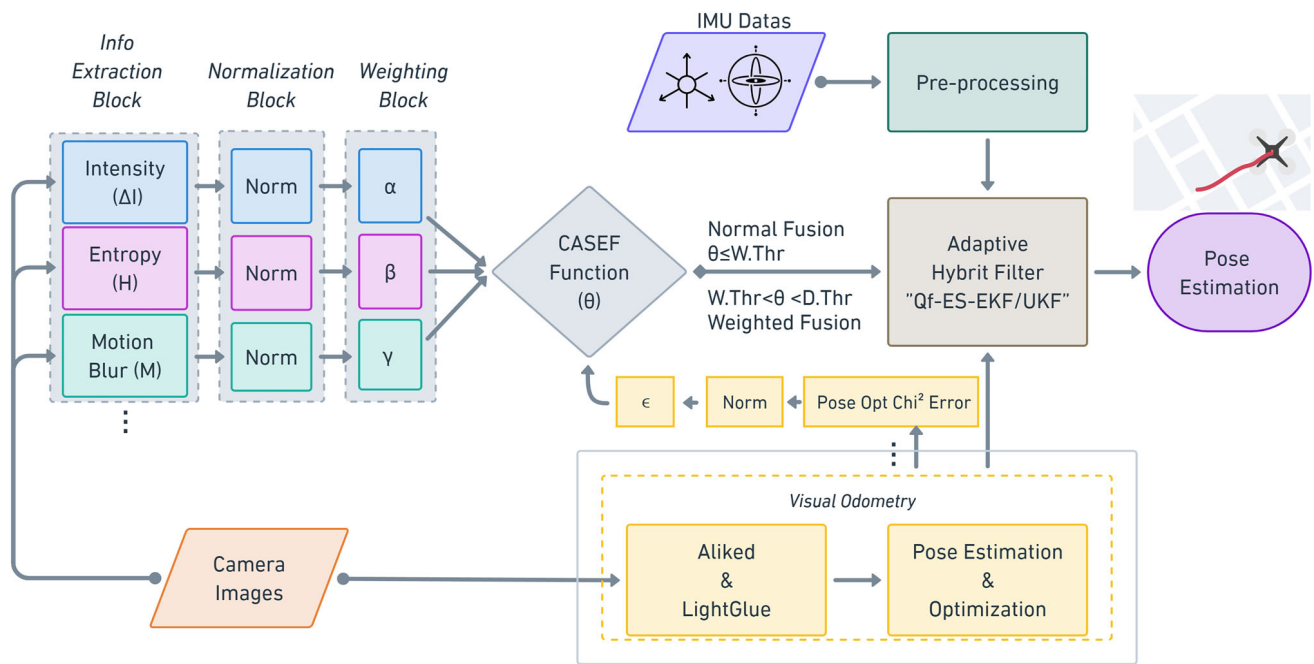

**Fig. 1** Block diagram of the proposed adaptive VIO approach: illustrating the interaction of components performing adaptive filtering and sensor reliability analysis

# 3 Proposed VIO System

The proposed system operates based on a loosely coupled sensor fusion architecture. This architecture incorporates a hybrid filtering approach named Quaternion-Focused Error-State EKF/UKF (Qf-ES-EKF/UKF), which is designed to process IMU data and is one of the core innovations of this study. The hybrid filter utilizes SUKF techniques to handle the highly non-linear aspects of orientation estimation, while leveraging the computational efficiency of the ESKF structure for other state components.

Another significant original contribution of the system is the adaptive covariance update mechanism used for integrating position and velocity measurements obtained from the visual odometry (VO) module into the filter. This mechanism instantaneously assesses the reliability of visual data. This assessment employs various image quality metrics and the novel Clipped Adaptive Saturation Exponential Function (CASEF) activation function, which will be detailed in Sect. 3.5.2. Based on the assessed confidence score, the measurement noise covariances are dynamically adjusted, thereby enhancing the overall accuracy and robustness of the system, especially in challenging scenarios where the quality of sensor data varies.

The overall architecture of the proposed VIO system and the interaction between its fundamental components are schematically illustrated in Fig. 1.

The system primarily consists of four main components: Qf-ES-EKF/UKF for Inertial Navigation, Visual Data Quality Control, Visual Odometry Module, and Sensor Fusion.

## 3.1 State and Error Representation

The proposed VIO system is based on an error-state formulation for state estimation, which is widely accepted in the literature [71]. In this approach, the true state of the system is modeled via a nominal state and an error state representing small deviations from this nominal state. The nominal state vector $\hat{x} \in \mathbb{R}^{16}$ and its corresponding error-state vector $\delta\mathbf{x} \in \mathbb{R}^{15}$ are defined in Eq. (1)

$$\hat{\mathbf{x}} = \begin{bmatrix} \hat{\mathbf{q}} \\ \hat{\mathbf{v}} \\ \hat{\mathbf{p}} \\ \hat{\mathbf{b}}_a \\ \hat{\mathbf{b}}_g \end{bmatrix}, \quad \delta\mathbf{x} = \begin{bmatrix} \delta\boldsymbol{\theta} \\ \delta\mathbf{v} \\ \delta\mathbf{p} \\ \delta\mathbf{b}_a \\ \delta\mathbf{b}_g \end{bmatrix}. \tag{1}$$

In Eq. (1), $\hat{\mathbf{q}} \in SO(3)$ is the unit quaternion representing the system's orientation. $\hat{\mathbf{p}}, \hat{\mathbf{v}} \in \mathbb{R}^3$ denotes the three-dimensional position and velocity of the system in the world reference frame, respectively. $\hat{\mathbf{b}}_a, \hat{\mathbf{b}}_g \in \mathbb{R}^3$ represents the estimated bias values for the accelerometer and gyroscope sensors, respectively.

In the error-state vector, $\delta\boldsymbol{\theta} \in \mathbb{R}^3$ represents a small rotational error around the nominal orientation $\hat{\mathbf{q}}$ and is considered an element of $so(3)$, the Lie algebra of $SO(3)$. The other error components $\delta\mathbf{v}, \delta\mathbf{p}, \delta\mathbf{b}_a, \delta\mathbf{b}_g \in \mathbb{R}^3$ represent the velocity and position errors in the world frame, and the standard Euclidean errors in accelerometer bias and gyroscope bias, respectively.

The relationship between the true state and the nominal state of the system, and the injection of the error state into the nominal state, are detailed in the standard ESKF literature [71]. This process is performed using $SO(3)$ group operations (e.g., quaternion multiplication and exponential map) for orientation and vector addition for other Euclidean states. After the filtering process is completed, the estimated error state $\delta\hat{\mathbf{x}}$ is injected into the nominal state, and the error state is reset to zero.

## 3.2 System Dynamics and Discretization

The evolution of a dynamic platform, such as a UAV, is defined by standard kinematic models that are crucial for tracking high-dynamics motion. These models rely on raw acceleration and angular velocity measurements from the on-board IMU, forming the core of state propagation in nearly all VIO systems designed for aerial robotics [26, 32]. This section addresses the nominal state and error-state dynamics and their discretization, which are fundamental to this process.

### 3.2.1 IMU Kinematics and Bias Modeling

The change over time of the nominal state vector $\hat{\mathbf{x}}$ (defined in Sect. 3.1) is modeled by the continuous-time kinematic equations given in Eq. (2)

$$\dot{\hat{\mathbf{q}}} = \frac{1}{2}\hat{\mathbf{q}} \otimes \begin{bmatrix} 0 \\ \boldsymbol{\omega}_m - \hat{\mathbf{b}}_g \end{bmatrix} \tag{2a}$$

$$\dot{\hat{\mathbf{p}}} = \hat{\mathbf{v}} \tag{2b}$$

$$\dot{\hat{\mathbf{v}}} = \mathbf{R}(\hat{\mathbf{q}})(\mathbf{a}_m - \hat{\mathbf{b}}_a) + \mathbf{g}. \tag{2c}$$

In Eq. (2), $\boldsymbol{\omega}_m$ and $\mathbf{a}_m$ denote the raw angular velocity and acceleration measured by the IMU. $\hat{\mathbf{b}}_g$ and $\hat{\mathbf{b}}_a$ represent the gyroscope and accelerometer bias estimates. $\mathbf{R}(\hat{\mathbf{q}})$ is the rotation matrix that transforms from the body frame to the world frame. Finally, $\mathbf{g}$ represents the gravity vector in the world frame. These nominal state dynamics are propagated using Euler or trapezoidal integration methods.

The change over time of accelerometer and gyroscope sensor biases ($\mathbf{b}_a, \mathbf{b}_g$) is modeled using first-order Gauss–Markov (GM) processes, which are widely used in the literature [71–73], to more realistically reflect their slowly varying and time-correlated nature. This model offers more stable bias estimates compared to the Random Walk model. The bias dynamics are expressed in Eq. (3)

$$\dot{\hat{\mathbf{b}}}_a = -\frac{1}{\tau_a}\hat{\mathbf{b}}_a + \mathbf{w}_{ba}, \qquad \dot{\hat{\mathbf{b}}}_g = -\frac{1}{\tau_g}\hat{\mathbf{b}}_g + \mathbf{w}_{bg}. \qquad (3)$$

Here, $\tau_a$ and $\tau_g$ represent the correlation time constants for the respective sensor biases, and $\mathbf{w}_{ba}$ and $\mathbf{w}_{bg}$ represent zero-mean, white Gaussian noise processes (bias random walk).

### 3.2.2 Error-State Dynamics and Discretization

The continuous-time dynamics of the error-state vector $\delta\mathbf{x}$ are linearized around the nominal state and biases. This error-state formulation, as detailed by Sola [71], is particularly critical for real-time applications on computationally constrained platforms like UAVs. It allows for efficient management of the integration drift inherent in MEMS sensors without requiring frequent re-linearization of the full, non-linear system state. The linearized dynamics are expressed as in Eq. (4)

$$\dot{\delta\mathbf{x}} = \mathbf{A}\delta\mathbf{x} + \mathbf{G}\mathbf{n}, \quad \text{where } \mathbf{n} \sim \mathcal{N}(\mathbf{0}, \mathbf{Q}_c). \qquad (4)$$

In this model, $\mathbf{A} \in \mathbb{R}^{15\times15}$ is the error-state dynamic matrix, $\mathbf{G} \in \mathbb{R}^{15\times12}$ is the noise distribution matrix, $\mathbf{n} \in \mathbb{R}^{12}$ is the system noise vector, and $\mathbf{Q}_c \in \mathbb{R}^{12\times12}$ is the continuous-time process noise covariance matrix. The matrices $\mathbf{A}$ and $\mathbf{G}$ depend on time-varying values such as the instantaneous nominal state ($\hat{\mathbf{q}}$), estimated biases ($\hat{\mathbf{b}}_a, \hat{\mathbf{b}}_g$), and measured IMU data ($\mathbf{a}_m, \boldsymbol{\omega}_m$). Their structures are shown in Eq. (5), following the standard ESKF literature [71]. The symbols $\mathbf{0}$ and $\mathbf{I}$ represent zero and identity matrices of appropriate dimensions (usually $3 \times 3$)

$$\mathbf{A} = \begin{bmatrix} -[\boldsymbol{\omega}_m - \hat{\mathbf{b}}_g]_\times & \mathbf{0} & \mathbf{0} & \mathbf{0} & -\mathbf{I} \\ -\mathbf{R}(\hat{\mathbf{q}})[\mathbf{a}_m - \hat{\mathbf{b}}_a]_\times & \mathbf{0} & \mathbf{0} & -\mathbf{R}(\hat{\mathbf{q}}) & \mathbf{0} \\ \mathbf{0} & \mathbf{I} & \mathbf{0} & \mathbf{0} & \mathbf{0} \\ \mathbf{0} & \mathbf{0} & \mathbf{0} & -\frac{1}{\tau_a}\mathbf{I} & \mathbf{0} \\ \mathbf{0} & \mathbf{0} & \mathbf{0} & \mathbf{0} & -\frac{1}{\tau_g}\mathbf{I} \end{bmatrix}, \quad \mathbf{G} = \begin{bmatrix} -\mathbf{I} & \mathbf{0} & \mathbf{0} & \mathbf{0} \\ \mathbf{0} & -\mathbf{R}(\hat{\mathbf{q}}) & \mathbf{0} & \mathbf{0} \\ \mathbf{0} & \mathbf{0} & \mathbf{0} & \mathbf{0} \\ \mathbf{0} & \mathbf{0} & \mathbf{I} & \mathbf{0} \\ \mathbf{0} & \mathbf{0} & \mathbf{0} & \mathbf{I} \end{bmatrix}. \qquad (5)$$

Here, $[\cdot]_\times$ is the operator that converts a vector into a skew-symmetric matrix. The continuous-time process noise covariance matrix is defined as $\mathbf{Q}_c = \text{diag}(\sigma_g^2\mathbf{I}, \sigma_a^2\mathbf{I}, \sigma_{wa}^2\mathbf{I}, \sigma_{wg}^2\mathbf{I})$, where the identity matrices $\mathbf{I}$ are of size $3 \times 3$.

For use in the filtering steps, these continuous-time dynamics must be discretized. In this study, we employ the Van Loan method [74], which provides an accurate discretization using the matrix exponential. This method is used to calculate the discrete-time state transition matrix $\boldsymbol{\Phi}_k$ and the process noise covariance $\mathbf{Q}_{d,k}$ over the IMU sampling period $\Delta t$. The method involves computing the exponential of an augmented matrix $\mathbf{M}(\Delta t)$, as defined in Eq. (6)

$$\mathbf{M}(\Delta t) = \begin{bmatrix} -\mathbf{A}\,\Delta t & \mathbf{G}\mathbf{Q}_c\mathbf{G}^\top\,\Delta t \\ \mathbf{0}_{15\times15} & \mathbf{A}^\top\,\Delta t \end{bmatrix} \in \mathbb{R}^{30\times30}, \qquad (6a)$$

$$\mathcal{E}(\Delta t) = \exp\big(\mathbf{M}(\Delta t)\big) = \begin{bmatrix} \mathcal{E}_{11} & \mathcal{E}_{12} \\ \mathcal{E}_{21} & \mathcal{E}_{22} \end{bmatrix}. \qquad (6b)$$

From the blocks of the resulting matrix $\mathcal{E}(\Delta t)$, the discrete-time matrices $\boldsymbol{\Phi}_k$ and $\mathbf{Q}_{d,k}$ are extracted as shown in Eq. (7)

$$\boldsymbol{\Phi}_k = \mathcal{E}_{22}^\top, \qquad (7a)$$

$$\mathbf{Q}_{d,k} = \boldsymbol{\Phi}_k\mathcal{E}_{12}^\top. \qquad (7b)$$

The matrices $\mathbf{A}$ and $\mathbf{G}$ within $\mathbf{M}(\Delta t)$ are evaluated at each time step using current state estimates. The resulting discrete matrices, $\mathbf{\Phi}_k$ and $\mathbf{Q}_{d,k}$, are then used in the propagation steps of our proposed hybrid filtering architecture.

## 3.3 State Propagation with Hybrid Qf-ES-EKF/UKF

The propagation stage of state estimation involves carrying the state vector and its uncertainty to the next time step using system dynamics and IMU measurements. The proposed Qf-ES-EKF/UKF filter adopts a hybrid strategy optimized for the different degree of non-linearity of the state vector components at this stage. This approach aims to provide a balanced solution between accuracy and computational efficiency, especially in VIO applications where the precision of orientation estimation is critical.

### 3.3.1 Rationale and Basic Approach of Hybrid Propagation

The dynamics of quaternions ($\hat{\mathbf{q}}$) representing system orientation are highly non-linear. While a standard ESKF struggles with this due to first-order linearization, a full SUKF, though more accurate, is computationally expensive. To address this trade-off, our proposed Qf-ES-EKF/UKF filter employs a hybrid propagation strategy that combines the strengths of both filters.

The process begins with a standard **ESKF propagation step**. In this step, the **final propagation** is performed for the more linearly behaved state variables, such as position, velocity, and biases. The same step also produces an initial ESKF-based estimate for the orientation. Subsequently, a targeted **SUKF refinement step** is applied exclusively to the non-linear orientation component. This second stage uses sigma points to re-propagate the orientation error, correcting for potential inaccuracies introduced by the ESKF's linearization and yielding a more precise final estimate for the system's rotation. This approach preserves the efficient and standard propagation of ESKF for other states while adding the focused accuracy of SUKF as an extra refinement layer only for orientation, thus avoiding the overhead of a full SUKF implementation.

### 3.3.2 Qf-ES-EKF/UKF Propagation Steps

The hybrid propagation process consists of two main steps:

#### *ESKF-Based Preliminary Propagation*

In the first step, using the state estimate $\hat{\mathbf{x}}_{k-1|k-1}$ and its covariance $\mathbf{P}_{k-1|k-1}$ at time $t_{k-1}$, a standard ESKF-based propagation is performed for the full nominal state $\hat{\mathbf{x}}$ and the error-state covariance $\mathbf{P}$. The nominal state is propagated with IMU measurements $\mathbf{u}_k$ according to Eq. (2) (based on the IMU kinematics in Sect. 3.2.1). Concurrently, the error-state covariance matrix is propagated using the discrete-time state transition matrix $\mathbf{\Phi}_k$ and process noise covariance $\mathbf{Q}_{d,k}$ defined in Sect. 3.2.2. As a result of this step, the preliminary propagated nominal state $\hat{\mathbf{x}}_{k|k-1}$ and error covariance $\mathbf{P}_{k|k-1}^{\text{ESKF}}$ for time $t_k$ are obtained.

#### *Refinement of Orientation Covariance with SUKF*

Following the ESKF-based propagation, an SUKF-based refinement step is applied to the orientation error covariance block ($\mathbf{P}_{\theta\theta,k|k-1}^{\text{ESKF}} = \mathbf{P}_{k|k-1}^{\text{ESKF}}[0:3, 0:3]$). The purpose of this step is to more accurately model the uncertainty of the highly non-linear orientation dynamics. In this step, sigma points $\boldsymbol{\chi}_{\delta\theta}^{(j)}$ are generated only for the 3-dimensional orientation error $\delta\boldsymbol{\theta}$ (defined in the body frame). This generation is performed using the Cholesky decomposition of $\mathbf{P}_{\theta\theta,k|k-1}^{\text{ESKF}}$ and standard SUT parameters [13]. These orientation error sigma points are converted into error quaternions $\delta\mathbf{q}(\boldsymbol{\chi}_{\delta\theta}^{(j)})$. The perturbed orientation sigma points $\mathbf{q}_{\text{inj}}^{(j)}$ are obtained by right-multiplying the nominal orientation $\hat{\mathbf{q}}_{k-1|k-1}$ at the previous time step $t_{k-1}$ (perturbing in the body frame) as shown in Eq. (8)

$$\mathbf{q}_{\text{inj}}^{(j)} = \hat{\mathbf{q}}_{k-1|k-1} \otimes \delta\mathbf{q}(\boldsymbol{\chi}_{\delta\theta}^{(j)}). \tag{8}$$

A critical feature of the proposed hybrid approach is computational efficiency and focused refinement. For this purpose, the injected orientations $\mathbf{q}_{\mathrm{inj}}^{(j)}$ are propagated to the next time step using the current IMU measurements $\mathbf{u}_k$ and the time interval $\Delta t_k$. This propagation is performed via the nominal IMU integration process in Eq. (2). During this process, other nominal state components (velocity, position, and biases at $t_{k-1}$) are kept constant. Each propagated orientation sigma point $\hat{\mathbf{q}}_{\mathrm{prop}}^{(j)}$ is retracted to a body-frame orientation error vector $\delta\boldsymbol{\theta}_{\mathrm{prop}}^{(j)}$ relative to the nominal orientation $\hat{\mathbf{q}}_{k|k-1}$. This operation is performed by referencing the nominal orientation $\hat{\mathbf{q}}_{k|k-1}$ at time $t_k$ obtained from the ESKF step and using the $SO(3)$ logarithmic map as shown in Eq. (9)

$$\delta\boldsymbol{\theta}_{\mathrm{prop}}^{(j)} = \mathrm{LogMapSO3}\left((\hat{\mathbf{q}}_{k|k-1})^{-1} \otimes \hat{\mathbf{q}}_{\mathrm{prop}}^{(j)}\right). \tag{9}$$

Using these propagated and retracted body-frame orientation error sigma points, the refined orientation error covariance $\mathbf{P}_{\theta\theta,k|k-1}^{\mathrm{SUKF}}$ is calculated according to Eq. (10) with standard SUT weights

$$\mathbf{P}_{\theta\theta,k|k-1}^{\mathrm{SUKF}} = \sum_{j=0}^{2n_h} W_c^{(j)} (\delta\boldsymbol{\theta}_{\mathrm{prop}}^{(j)} - \hat{\delta\boldsymbol{\theta}}_{k|k-1}^{\mathrm{SUKF}})(\delta\boldsymbol{\theta}_{\mathrm{prop}}^{(j)} - \hat{\delta\boldsymbol{\theta}}_{k|k-1}^{\mathrm{SUKF}})^T. \tag{10}$$

Here, $\hat{\delta\boldsymbol{\theta}}_{k|k-1}^{\mathrm{SUKF}}$ is the weighted average of the propagated sigma points, which is theoretically zero due to symmetric sigma point selection.

### Hybrid Covariance Update

Finally, the orientation block (submatrix with indices $[0:3, 0:3]$) of the full error covariance matrix $\mathbf{P}_{k|k-1}^{\mathrm{ESKF}}$, propagated by ESKF, is updated. This update is performed using the SUKF-based refined orientation covariance $\mathbf{P}_{\theta\theta,k|k-1}^{\mathrm{SUKF}}$, as shown in Eq. (11). The cross-correlation terms between orientation and other states, as well as the covariance blocks of other states themselves, are preserved with their values from the ESKF propagation. This selective update focuses the refinement effect of SUKF specifically on modeling orientation uncertainty, while simultaneously maintaining the efficiency of ESKF for other states. Finally, the top-left $3 \times 3$ block of the full error covariance matrix $\mathbf{P}_{k|k-1}^{\mathrm{ESKF}}$, propagated by ESKF, corresponding to the orientation error covariance, is updated using the SUKF-based refined orientation covariance $\mathbf{P}_{\theta\theta,k|k-1}^{\mathrm{SUKF}}$. This updated orientation covariance block is denoted as $\mathbf{P}_{\theta\theta,k|k-1}$ and calculated as shown in Eq. (11).

$$\mathbf{P}_{\theta\theta,k|k-1} = \frac{1}{2}\left(\mathbf{P}_{\theta\theta,k|k-1}^{\mathrm{SUKF}} + (\mathbf{P}_{\theta\theta,k|k-1}^{\mathrm{SUKF}})^T\right). \tag{11}$$

This updated block $\mathbf{P}_{\theta\theta,k|k-1}$ replaces the corresponding $3 \times 3$ top-left block of the full $\mathbf{P}_{k|k-1}$ matrix. The cross-correlation terms between orientation and other states, as well as the covariance blocks of other states themselves, are preserved with their original values from the $\mathbf{P}_{k|k-1}^{\mathrm{ESKF}}$ matrix obtained from ESKF propagation. This selective update focuses the refinement effect of SUKF specifically on modeling orientation uncertainty, while simultaneously maintaining the efficiency of ESKF for other states.

### 3.3.3 Computational Complexity Analysis

The propagation step of a standard ESKF has a complexity of $\mathcal{O}(n^2)$ for an $n = 15$ dimensional state. A full SUKF implementation can be on the order of $\mathcal{O}(n^3)$ (with Cholesky decomposition and sigma point propagation). In the proposed hybrid Qf-ES-EKF/UKF approach, after the full ESKF propagation is performed, the SUKF step is applied only for the $n_h = 3$ dimensional orientation error. The complexity of this SUKF step is approximately $\mathcal{O}(n_h^3)$. The part of ESKF affecting the remaining $n - n_h = 12$ dimensions remains $\mathcal{O}((n - n_h)^2)$, and the cross terms also remain of a similar order. Therefore, the overall complexity of the hybrid approach can be roughly expressed as $\mathcal{O}(n_h^3 + (n - n_h)^2)$. This is significantly lower compared to a full-state SUKF implementation. This situation offers a significant advantage in achieving the critical accuracy-efficiency balance for real-time VIO applications.

## 3.4 Measurement Update Strategy

Following the state propagation step performed with IMU data, the system's state estimate and uncertainty are updated using information from various external measurement sources. The proposed filter structure incorporates three fundamental measurement update modules, which are applied sequentially within the standard EKF framework [75]: Zero Velocity Update (ZUPT), Gravity Alignment, and Visual Position-Velocity Update. Each update follows the standard EKF steps of calculating the innovation, Kalman gain, and performing the state and covariance correction using its respective measurement model $h(\cdot)$, measurement Jacobian $\mathbf{H}$, and measurement noise covariance $\mathbf{R}$.

It should be noted that, in the proposed hybrid Qf-ES-EKF/UKF architecture, while SUKF principles are applied for orientation error in the state propagation (prediction) step as detailed in Sect. 3.3, a standard ESKF logic is used for all error states (including orientation) in the measurement update steps described here. This approach aims to improve accuracy while preserving computational efficiency, as the measurement models are generally linear or easily linearizable, and the main system non-linearities stem from the IMU dynamics.

The first update mechanism is the **Zero Velocity Update (ZUPT)**. When the system is detected to be static (i.e., the standard deviations of IMU accelerometer and gyroscope measurements fall below certain thresholds), the velocity in the body reference frame is assumed to be zero ($\mathbf{z}_{\text{ZUPT}} = \mathbf{0}$). The measurement model is defined by Eq. (12a), and the corresponding Jacobian matrix $\mathbf{H}_{\text{ZUPT}}$ is presented in Eq. (12b)

$$h_{\text{ZUPT}}(\hat{\mathbf{x}}) = \mathbf{R}(\hat{\mathbf{q}})^T \hat{\mathbf{v}}. \tag{12a}$$

$$\mathbf{H}_{\text{ZUPT}} = \left[ [\mathbf{R}(\hat{\mathbf{q}})^T \hat{\mathbf{v}}]_\times \ \mathbf{R}(\hat{\mathbf{q}})^T \ \mathbf{0}_{3\times9} \right]. \tag{12b}$$

This update corrects the orientation and velocity estimates, and the nominal velocity estimate $\hat{\mathbf{v}}$ is directly reset to zero. The measurement noise covariance is defined as $\mathbf{R}_{\text{ZUPT}}$.

Second, a **Gravity Alignment Update** is performed in quasi-static situations where the system's acceleration is negligibly low. In such cases, the IMU accelerometer readings ($\mathbf{a}_m$) predominantly reflect the gravity vector. This information is used to correct the system's orientation (especially roll and pitch angles) and the accelerometer bias $\hat{\mathbf{b}}_a$. The measurement is taken as $\mathbf{z}_g = \mathbf{a}_m - \hat{\mathbf{b}}_a$, with the measurement model given in Eq. (13) and the corresponding Jacobian matrix $\mathbf{H}_g$ in Eq. (14)

$$h_g(\hat{\mathbf{x}}) = \mathbf{R}(\hat{\mathbf{q}})^T \mathbf{g}. \tag{13}$$

$$\mathbf{H}_g = \left[ [\mathbf{R}(\hat{\mathbf{q}})^T \mathbf{g}]_\times \ \mathbf{0}_{3\times3} \ \mathbf{0}_{3\times3} \ -\mathbf{I}_{3\times3} \ \mathbf{0}_{3\times3} \right]. \tag{14}$$

The measurement noise covariance is determined as $\mathbf{R}_{\text{acc}} = \sigma_a^2 \mathbf{I}_{3\times3}$ using the accelerometer noise standard deviation $\sigma_a$.

Finally, the **Visual Position and Velocity Update** integrates measurements from the external VO module. In this loosely coupled scheme, the system's state estimation is updated with position $\mathbf{p}_{\text{vis}}$ and velocity $\mathbf{v}_{\text{vis}}$ data. The filter's estimated position $\hat{\mathbf{p}}$ and velocity $\hat{\mathbf{v}}$ are compared with the VO outputs through the linear measurement model defined in Eq. (15) and its corresponding Jacobian matrix $\mathbf{H}_{\text{VIS}}$ in Eq. (16)

$$\mathbf{z}_{\text{VIS}} = \begin{bmatrix} \mathbf{p}_{\text{vis}} \\ \mathbf{v}_{\text{vis}} \end{bmatrix}, \qquad h_{\text{VIS}}(\hat{\mathbf{x}}) = \begin{bmatrix} \hat{\mathbf{p}} \\ \hat{\mathbf{v}} \end{bmatrix}. \tag{15}$$

$$\mathbf{H}_{\text{VIS}} = \begin{bmatrix} \mathbf{0}_{3\times3} & \mathbf{0}_{3\times3} & \mathbf{I}_{3\times3} & \mathbf{0}_{3\times6} \\ \mathbf{0}_{3\times3} & \mathbf{I}_{3\times3} & \mathbf{0}_{3\times3} & \mathbf{0}_{3\times6} \end{bmatrix}. \tag{16}$$

The noise covariance matrix for these visual measurements is $\mathbf{R}_{\text{VIS}} = \text{diag}(\sigma_p^2 \mathbf{I}_{3\times3}, \sigma_v^2 \mathbf{I}_{3\times3})$. As a key contribution of this study, the variances $\sigma_p^2$ and $\sigma_v^2$ are adaptively updated based on visual data quality, a process detailed in Sect. 3.5.

## 3.5 Adaptive Update Mechanism

The adaptive update mechanism performs a weighted fusion by quantitatively assessing the reliability of the data from each sensor. The first and most critical step in this process is to analyze the quality of the incoming visual data.

### 3.5.1 Visual Data Quality Analysis

Upon reviewing numerous visual datasets in the literature, it is observed that sudden light changes, low texture density, and rapid camera movements are among the primary factors that directly and negatively impact the performance of visual odometry systems. Such challenges have become characteristic of demanding scenarios and have been reported as fundamental factors limiting the reliability of visual estimation systems [76–78]. In this context, various image processing-based methods for detecting and grading these visual difficulties have been widely proposed in the literature. For assessing illumination conditions in the scene, histogram-based analyses [79, 80], local contrast measurements [81, 82], and frequency domain analyses [83, 84] are frequently used. These methods allow for the numerical measurement of the brightness level and variability in the scene through the intensity distribution and frequency components of images. In the analysis of texture density, methods, such as local entropy maps [85], multi-scale entropy analysis [86], and joint entropy between color channels [87], are prominent. These techniques measure the amount of structural information in the image, quantitatively representing the complexity of surface textures. For the assessment of motion blur, spectral analysis approaches based on Fourier transform [88], spatial gradient measurements based on edge sharpness [89], and deep learning-assisted blur estimation methods [90] are commonly used. These methods offer a wide range of solutions, including both classical signal processing techniques and modern data-driven learning approaches.

Some of these methods proposed in the literature are notable not only for their theoretical accuracy but also for their computational efficiency. For example, *intensity*-based measurements based on average pixel intensity have a time complexity of $O(N)$, proportional to the total number of pixels $N$. Similarly, methods based on *Shannon entropy* offer a time complexity at the level of $O(N + 256)$ when histogram calculation and entropy summation are considered together. The *Laplacian variance*, frequently used in determining motion blur, also operates at the $O(N)$ level via pixel-based edge response.

All of these methods carry high potential in terms of application, as they can be integrated into real-time visual odometry systems. Especially, these indicators, which provide high discriminative capability despite low computational load, stand out as structures that can be effectively used for reliability analysis in complex scene conditions.

The static illumination level in the scene has been successfully monitored with the *intensity* metric, while temporal light changes have been monitored via the *delta intensity* metric. However, it is clearly visible in Fig. 4 that the values obtained with the *Shannon entropy* method, which has a high capacity to represent the structural information content in the scene, show a high correlation with *static intensity* measurements. Evaluating these findings, it is understood that there is no need for separate system integration of these two highly correlated metrics. As a result of analyses conducted on five different "Machine Hall" sequences in the EuRoC MAV dataset, it was observed in comparative analyses that the most consistent measures representing the illumination level were *Shannon entropy* for the static component and *delta intensity* for temporal change. These findings show that the metrics presented in Fig. 5 consistently align with the unique metadata of each sequence. Equation (17) presents the formulation for these metrics

$$I = \frac{1}{N}\sum_{i=1}^{N} p_i, \quad intensity_{\mathcal{N}} = \frac{I}{I_{\max} - I_{\min}}, \quad \Delta intensity_{\mathcal{N}} = intensity_{\mathcal{N}}\big|_{t-1}^{t},$$

$$H = -\sum_{i=0}^{255} p_i \log_2(p_i), \quad entropy_{\mathcal{N}} = \frac{H}{H_{\max} - H_{\min}}, \tag{17}$$

$$MB = \mathrm{var}(\nabla^2 I), \quad blur_{\mathcal{N}} = \frac{MB}{MB_{\max} - MB_{\min}}, \quad \Delta blur_{\mathcal{N}} = blur_{\mathcal{N}}\big|_{t-1}^{t},$$

where

- $p_i$: intensity value of the $i$-th pixel (in the range [0, 256]),
- $N$: total number of pixels (image resolution),
- $(\cdot)_{\max}$, $(\cdot)_{\min}$: maximum and minimum values referenced during normalization,
- $(\cdot)_{\mathcal{N}}$: indicates the normalized version of the metric,
- $\Delta(\cdot)_{\mathcal{N}}$: normalized metric difference between two consecutive frames (e.g., $\Delta blur_{\mathcal{N}} = blur_{\mathcal{N}}\big|_{t-1}^{t}$),
- $\mathrm{var}(\nabla^2 I)$: variance of the *Laplacian operator*; used for motion blur estimation,
- $H = -\sum_{i=0}^{255} p_i \log_2(p_i)$: *Shannon entropy* calculation; $p_i$ is the gray-level probability.

Figure 2 shows scenes with extreme values for each metric, whose calculation method is detailed in Eq. (17).

It has been observed that the proposed measurement methods can successfully detect challenges such as illumination changes and low texture content. However, while motion blur is expected to be triggered by velocity measurements in dynamic scenes, it was found to reach its highest level in static scenes.

Although the variance of the *Laplacian operator* is not limited to identifying only high-speed scenes, it should be considered that the primary factor causing difficulty in image matching is blur, not speed. Therefore, the contribution of the relevant metric was included in the analysis for evaluation.

It is possible to use additional metrics to evaluate the quality of visual inferences with probabilistic approaches. In this context, in addition to indicators such as *projection error* that can quantitatively measure inference performance, metrics such as *Pose Optimization $\chi^2$ Error*, which showed high correlation with the success of visual inferences in our experimental analyses, and the number of *keyframes* culled by the system were also used in the same way. Since these indicators are obtained directly during the visual inference process, unlike analyses based on prior information, they are discussed in more detail in the relevant visual odometry section.

### 3.5.2 Confidence Score Calculation

In traditional loosely coupled sensor fusion methods, the visual position-velocity update is typically triggered solely based on the availability of new visual measurements, without considering their underlying quality or reliability. For example, if a minimum of four matching feature points are not obtained from consecutive image pairs, the visual update is skipped, and only IMU propagation continues. This approach disregards the accuracy and quality of visual measurements, giving equal importance to all measurements, which prevents the measurement noise from being appropriately reflected in the filter. In the proposed method, to overcome this limitation, a reliability score, $\theta \in [0, 1]$, is calculated to quantitatively assess the reliability of each visual measurement against a **Weighting Threshold** ($W_{thr}$). If $\theta$ exceeds this threshold, the visual data are considered suspect, and its corresponding measurement noise covariance is increased proportionally to the degree of unreliability. The conceptual limit at which visual data are deemed completely unreliable corresponds to the point where our CASEF activation function naturally saturates at its maximum value of 1. At this point, the measurement covariance reaches its predefined maximum, effectively nullifying the impact of the visual update on the state estimate. This design choice elegantly handles the "deadline" scenario without the need for a separate, hard-coded deadline threshold, as the saturation characteristic of the CASEF function provides an inherent and smooth transition to maximum uncertainty. This mechanism prevents the assimilation of potentially corrupt visual data, aiming to enhance overall system performance. If the $\theta$ value is within the reliable range ($\theta \le W_{thr}$), the system operates like a traditional loosely coupled sensor fusion system, trusting the visual measurements with minimum covariance.

**Fig. 2** Scenes with maximum metric values in the EuRoC MAV-Machine Hall dataset

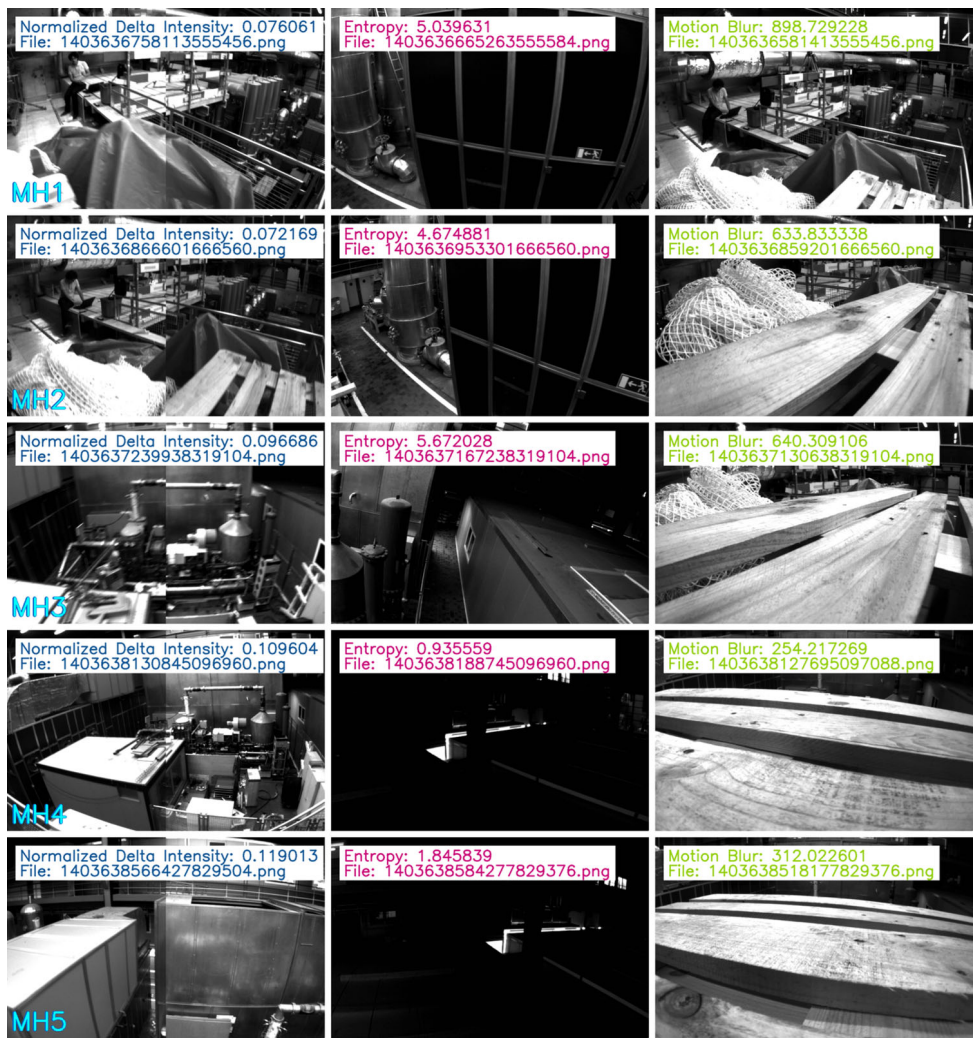

The confidence metrics $\theta_p$ and $\theta_v$, which evaluate image quality and reliability, are calculated separately based on static and dynamic criteria using Eq. (18a) and Eq. (18b). In Eq. (18a), the normalized *Shannon entropy* ($entropy_{\mathcal{N}}$), since high values indicate good image quality, is inverted as ($1 - entropy_{\mathcal{N}}$) to be used as a "badness" or "unreliability" metric. Thus, the $\theta_p$ value is always determined by the metric corresponding to the worst-case scenario. Here, $\theta_p$ represents static criteria, and $\theta_v$ represents dynamic criteria

$$\theta_p = f\Big(\max\big\{(1 - entropy_{\mathcal{N}}),\ blur_{\mathcal{N}},\ \chi^2_{pose_{\mathcal{N}}},\ keyf^c_{\mathcal{N}}\big\}\Big) \tag{18a}$$

$$\theta_v = f\Big(\max\big\{\alpha\ \Delta intensity_{\mathcal{N}},\ \beta\ \Delta blur_{\mathcal{N}},\ \gamma\ \Delta \chi^2_{pose_{\mathcal{N}}},\ \zeta\ \Delta keyf^c_{\mathcal{N}}\big\}\Big), \tag{18b}$$

where

- $(1 - entropy_\mathcal{N})$: The inverted value of the normalized *Shannon entropy*. Since high entropy indicates good image quality, this value is transformed into $1 - entropy_\mathcal{N}$ to be used as a "badness" or "unreliability" metric. Thus, images with low entropy (i.e., high $(1 - entropy_\mathcal{N})$ value) contribute to a higher unreliability score.
- $\Delta\chi^2_{pose_\mathcal{N}}$: Represents the normalized change in the $\chi^2$ (chi-squared) error value obtained from pose optimization between consecutive visual measurements. (This metric will be explained in detail in the visual odometry section.)
- $\Delta keyf^c_\mathcal{N}$: Represents the normalized value of the decrease in the number of tracked keyframes. (This metric will be explained in detail in the visual odometry section.)
- $\alpha, \beta, \gamma, \zeta$: Weighting factors that determine the contribution of the criteria to the confidence metric.
- $f$: Activation function (e.g., CASEF, ReLU, and Sigmoid), which limits the calculated value to the range [0, 1] or emphasizes values above a certain threshold.
- max(): Selects the highest value among the given criteria, ensuring that the confidence value is based on the most prominent feature.

This score is obtained from a combination of quality metrics, such as image entropy, pixel intensity, and motion blur. The measurement noise covariance $\mathbf{R}_{VIS}$ is initially defined with constant values $[\sigma^2_p, \sigma^2_v]$. However, these values are adaptively updated between minimum and maximum values determined according to the system's characteristics. Metrics representing the static properties of the scene, for example, the illumination value, affect the position covariance $\sigma^2_p$. The temporal change of this metric, i.e., the intensity difference resulting from instantaneous light change, affects the velocity covariance $\sigma^2_v$. This approach is consistent with the principle that velocity is the temporal derivative of position. Thus, the impact of measurements with high error potential on the system is reduced, significantly improving estimation accuracy and consistency.

### Activation Function

The confidence metric $\theta$ is the fundamental parameter that determines which sensor data the system will trust more. However, changes in $\theta$ within certain ranges, especially causing a sudden decrease in confidence in visual data, can negatively affect system performance. For example, the reliability of a visual inference based on only 4 matching features cannot be represented by the same linear probability model as a visual inference based on 20 matching features. Accordingly, a linear change in confidence in the $\theta$ value may not accurately reflect adaptive system behavior.

To overcome this problem, various activation functions with saturation characteristics have been proposed in the literature. Among these functions, functions like *Quartic Unit step* and *Exponential Surge* can model the confidence distribution between sensors more realistically. However, these functions cannot guarantee optimal performance in all system states. Therefore, the CASEF proposed in our study can adjust the plateau value of the activation function with different $s$ parameters; it produces a result of 0 for negative values and 1 for values greater than 1. Thus, CASEF offers both a flexible and controlled structure in adaptive confidence modeling. The mathematical model of the CASEF function is presented in Eq. (19)

$$\text{CASEF}(x; s) = \frac{\exp(s \cdot \text{clip}(x, 0.0, 1.0)) - 1}{\exp(s) - 1}. \tag{19}$$

Here, $x$ represents the input value, and the $s$ parameter adjusts the plateau value of the function. The function ensures that $x$ is clipped between 0 and 1 (*clip*) and applies an exponential increase after this clipping operation. A comparison of the CASEF function with classical static activation functions is presented in Fig. 3.

This proposed probabilistic weighting approach has the potential to overcome the limitations of traditional methods and provide more robust and adaptive VIO performance in varying environmental conditions. The effectiveness of this approach is supported by the experimental results presented in Sect. 5.

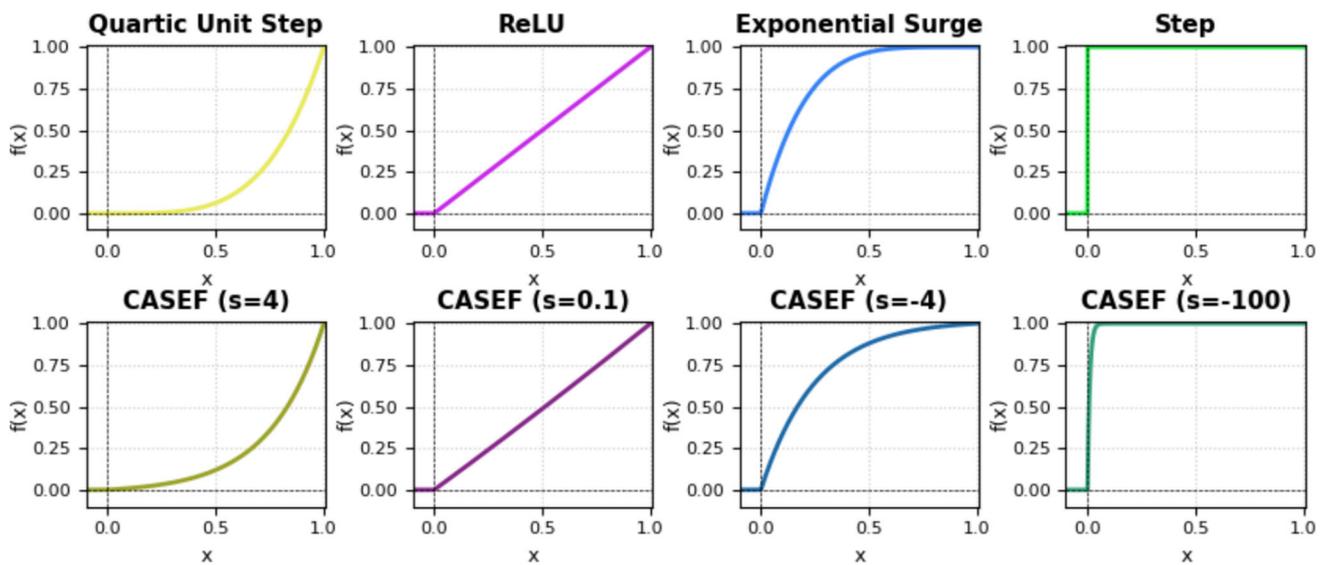

**Fig. 3** Comparison of the CASEF function with classical static activation functions

## 3.6 State Estimation with Visual Odometry

In the proposed system, a lightweight VO architecture based on PySLAM [61] is used for real-time camera pose estimation. This architecture extracts features from stereo image pairs using ALIKED [56]. It then integrates matching with LightGlue [8], depth calculation with Semi-Global Block Matching (SGBM) [91], and sliding window optimization to provide high-frequency pose estimation. Due to low-latency operational requirements, the system does not include global optimization steps such as full bundle adjustment (BA) and loop closure. Instead, it adopts an incremental approach.

For each stereo pair, 512 2D feature points and descriptors are first extracted from the left and right images using the ALIKED algorithm. LightGlue matches these points under epipolar geometry constraints to create a disparity map. 3D point clouds are triangulated using the semi-global block matching (SGBM) method for depth information calculation. During this process, geometric consistency is ensured by RANSAC-based outlier rejection.

Camera poses are estimated through PnP (Perspective-n-Point) optimization between consecutive frames. PySLAM's existing graph optimization module has been extended. With this extension, a sliding window approach using Lie algebra (SE(3)) parameterization is implemented. Poses and associated 3D points within a window consisting of the last N keyframes are optimized using the g2o library to minimize reprojection errors [92]. Keyframe selection is performed dynamically based on parallax thresholds and common visibility analysis.

To maintain local consistency, a limited bundle adjustment (Local Bundle Adjustment, Local BA) is performed on the 3D points associated with keyframes within the active window. This optimization covers only the last few frames and the points associated with them in the covariance graph, thus keeping the computational load under control. Excessively old frames are removed from the window to optimize memory usage.

Some fundamental statistical metrics have been recorded in real time to quantitatively monitor system performance. Some of these are as follows:

- `num_matched_kps`: Number of features extracted by ALIKED and matched by LightGlue.
- `num_inliers`: Number of geometrically consistent points remaining after RANSAC.
- `num_kf_ref_tracked_points`: % tracking rate of points in the reference keyframe; < 60% triggers a new keyframe.
- `descriptor_distance_sigma`: Standard deviation of LightGlue descriptor distances. > 0.25 indicates a risk of outlier matches.

- `last_num_static_stereo_points`: Number of 3D points considered static after calibration. Reflects the geometric stability of the environment.
- `last_num_triangulated_points`: Number of points triangulated from the last stereo pair.
- `last_num_fused_points`: Number of super-points considered identical from multiple viewpoints.
- `last_num_culled_keyframes`: Number of unnecessary keyframes removed from the covariance graph. Reflects memory optimization.
- `last_num_culled_points`: Number of 3D points culled due to geometric inconsistency. An indicator of map quality.
- `projection_error`: Average reprojection error; > 2.5 pixels indicates a risk of calibration loss.

The *Pose Optimization* $\chi^2$ *Error* and *Projection Error* metrics, used to evaluate the quality of pose estimation in visual odometry, both rely on reprojection errors but measure different aspects. *Pose Optimization* $\chi^2$ *Error* is a statistical metric obtained at the end of optimization processes. It is calculated by squaring the reprojection errors, scaling them by the associated measurement uncertainty, and then summing or averaging them over all points. In contrast, *Projection Error* is a geometric metric that measures the average pixel distance between the 2D feature points observed and the projection of the 3D points (obtained using the optimized pose) onto the image plane. In this calculation, scaling by measurement variance is not performed; only the magnitude of the deviation in the pixel plane is directly expressed. In summary, while *Pose Optimization* $\chi^2$ *Error* statistically evaluates the conformity of the data to the noise model, *Projection Error* geometrically indicates the magnitude of the direct pixel deviation.

The algorithmic structure of the adaptive covariance module is presented in detail in Algorithm 1, comprehensively explaining the calculation of relevant confidence metrics and the measurement covariance update steps.

## 4 Experimental Results and Discussion

This section presents a comprehensive experimental evaluation designed to validate the performance and demonstrate the key contributions of the proposed adaptive VIO system. As the foundation of the system's adaptive structure is its ability to instantaneously assess the reliability of data from visual odometry, the experimental analysis commences with the identification of the most suitable metrics to represent this confidence. The selection of these metrics is based on a quantitative analysis of the correlation between each candidate metric and the final Absolute Trajectory Error (ATE), aimed at determining the extent to which they reflect overall system error. This process, substantiated by the correlation matrix presented in Fig. 4, ensures the integration of the most indicative metrics into the system.

Subsequent to the identification of effective metrics, a critical next step involves finding the optimal set of hyperparameters ($\alpha$, $\beta$, $\gamma$, $\zeta$, $W_{thr}$, $s$) that govern their influence on the adaptive filter. To this end, a two-stage optimization strategy, detailed in Fig. 6, was employed. This strategy incorporates a Genetic Algorithm (GA) targeting a global optimum by accounting for complex inter-parameter effects, ensuring that a single, robust set of parameters is obtained to maintain consistency across all experiments.

Once the experimental framework was thus established and the parameters were meticulously optimized, the system's performance was examined along two primary axes. First, the core advantages of the proposed Qf-ES-EKF/UKF filter architecture are demonstrated at the component level, through a comparative analysis of its orientation estimation accuracy and computational efficiency against the standard ESKF and SUKF methods. Finally, the holistic contribution of the adaptive covariance mechanism is proven by evaluating the end-to-end performance of the complete Adaptive Qf-ES-EKF/UKF system, quantifying its superior positional accuracy against a non-adaptive baseline, particularly in challenging scenarios.

---

**Algorithm 1** Adaptive Covariance Update Mechanism

---

**Require:**
   *Visual Metrics (at time t):*
     - Static: *Shannon entropy, Laplacian variance* (blur), *Pose Optimization* $\chi^2$ *Error, Culled Keyframes.*
     - Dynamic (delta values): $\Delta$*intensity*, $\Delta$*blur*, $\Delta$*Pose Optimization* $\chi^2$ *Error*, $\Delta$*Culled Keyframes.*
**Require:**
   *User-Defined Parameters:*
     - Thresholds: Weighting Threshold ($W_{thr}$), Deadline Threshold ($D_{thr}$).
     - Scaling factors: $\alpha, \beta, \gamma, \zeta$.
     - Covariance bounds: $Min\_Cov_p, Max\_Cov_p, Min\_Cov_v, Max\_Cov_v$.
**Ensure:** Updated visual measurement covariance matrix $\mathbf{R}_{vis}$.

1: **for** each new visual measurement instance $\mathbf{z}_{vis}$ at time $t$ **do**
2:     *// Normalize all input metrics to a [0, 1] range, denoted by $(\cdot)_{\mathcal{N}}$*
3:     **Assess Static Reliability for Position Covariance ($\sigma_p$)**
4:       $u_p \leftarrow \max\left(1 - entropy_{\mathcal{N}}, \ blur_{\mathcal{N}}, \ PoseOpt\chi^2Error_{\mathcal{N}}, \ CulledKeyframes_{\mathcal{N}}\right)$
5:       $\theta_p \leftarrow$ CASEF$(u_p, s)$               ▷ Apply CASEF activation with parameter $s$
6:       **if** $\theta_p > D_{thr}$ **then**
7:         $\sigma_p \leftarrow Max\_Cov_p$           ▷ Visual data unreliable, maximize position uncertainty
8:       **else if** $\theta_p > W_{thr}$ **then**
9:         $\sigma_p \leftarrow Min\_Cov_p + \theta_p \cdot (Max\_Cov_p - Min\_Cov_p)$        ▷ Weight uncertainty
10:      **else**
11:        $\sigma_p \leftarrow Min\_Cov_p$          ▷ Visual data reliable, minimize position uncertainty
12:      **end if**
13:     **Assess Dynamic Reliability for Velocity Covariance ($\sigma_v$)**
14:       $u_v \leftarrow \max\left(\alpha \ \Delta intensity_{\mathcal{N}}, \ \beta \ \Delta blur_{\mathcal{N}}, \ \gamma \ \Delta PoseOpt\chi^2Error_{\mathcal{N}}, \ \zeta \ \Delta CulledKeyframes_{\mathcal{N}}\right)$
15:       $\theta_v \leftarrow$ CASEF$(u_v, s)$               ▷ Apply CASEF activation with parameter $s$
16:       **if** $\theta_v > D_{thr}$ **then**
17:         $\sigma_v \leftarrow Max\_Cov_v$           ▷ Visual data unreliable, maximize velocity uncertainty
18:       **else if** $\theta_v > W_{thr}$ **then**
19:         $\sigma_v \leftarrow Min\_Cov_v + \theta_v \cdot (Max\_Cov_v - Min\_Cov_v)$        ▷ Weight uncertainty
20:      **else**
21:        $\sigma_v \leftarrow Min\_Cov_v$          ▷ Visual data reliable, minimize velocity uncertainty
22:      **end if**
23:     **Construct Final Measurement Covariance Matrix**
24:       $\mathbf{R}_{vis} \leftarrow \text{diag}\left(\sigma_p^2 \cdot \mathbf{I}_{3\times3}, \sigma_v^2 \cdot \mathbf{I}_{3\times3}\right)$
25:     **Perform Measurement Update**
26:       Update filter with $\mathbf{z}_{vis}$ using the adaptive covariance $\mathbf{R}_{vis}$.
27: **end for**

---

## 4.1 Evaluation Methodology

In this study, the performance of the proposed VIO system was evaluated using positional and orientational accuracy metrics. Before all error analyses, the estimated trajectory and the reference (ground truth) trajectory were subjected to an optimal alignment process in $SE(3)$ space. This alignment was performed using the evo Python package [93]. The implementation of our system is publicly available for reproducibility.[1]

### 4.1.1 Positional Accuracy Metrics

The positional accuracy of the system was primarily measured by the ATE metric. ATE is obtained by calculating the RMSE of the absolute positional differences between the aligned estimated trajectory and the reference trajectory. The general ATE metric is shown in Equation (20a), and the individual positional errors ($E_a$) for the $x$, $y$, $z$ axes

---

[1] The source code for this research is publicly available in the dedicated GitHub repository.

**Table 1**  The EuRoC MAV dataset machine hall specifics [76]

| Sequence | Length / Duration | Avg. Speed / Angular Speed | Metadata |
| --- | --- | --- | --- |
| MH01 easy | 80.6 m / 182 s | 0.44 m s$^{-1}$ / 0.22 rad s$^{-1}$ | Good texture, bright scene |
| MH02 easy | 73.5 m / 150 s | 0.49 m s$^{-1}$ / 0.21 rad s$^{-1}$ | Good texture, bright scene |
| MH03 medium | 130.9 m / 132 s | 0.99 m s$^{-1}$ / 0.29 rad s$^{-1}$ | Fast motion, bright scene |
| MH04 hard | 91.7 m / 99 s | 0.93 m s$^{-1}$ / 0.24 rad s$^{-1}$ | Fast motion, dark scene |
| MH05 hard | 97.6 m / 111 s | 0.88 m s$^{-1}$ / 0.21 rad s$^{-1}$ | Fast motion, dark scene |

are calculated, as shown in Equation (20b)

$$\text{ATE} = \sqrt{\frac{1}{N} \sum_{i=1}^{N} \|\mathbf{p}_{i,\text{est}} - \mathbf{p}_{i,\text{gt}}\|_2^2},$$ (20a)

$$E_a = \sqrt{\frac{1}{N} \sum_{i=1}^{N} \left(a_{i,\text{est}} - a_{i,\text{gt}}\right)^2}, \quad a \in \{x, y, z\},$$ (20b)

where $N$ is the total number of time steps, $\mathbf{p}_{\text{est}}$ and $\mathbf{p}_{\text{gt}}$ are the estimated and ground truth 3D positions respectively, $a_{\text{est}}$ and $a_{\text{gt}}$ are the coordinate values on the respective axes, and $\|\cdot\|_2$ represents the Euclidean norm.

### 4.1.2 Orientational Accuracy Metrics

Orientation estimation performance was evaluated based on two fundamental metrics:

1. **Euler Angle Errors:** For each Euler angle (roll, pitch, yaw), the RMSE of the differences between the estimated and reference values was calculated in degrees.
2. **Overall Quaternion Orientation Error:** The RMSE of the angular magnitude ($\Delta\theta_{\text{rot}}$) of the relative rotation between the estimated quaternion $\mathbf{q}_{\text{est}}$ and the true quaternion $\mathbf{q}_{\text{gt}}$, as expressed in Eq. (21), was calculated

$$\Delta\theta_{\text{rot}}(\mathbf{q}_{\text{est}}, \mathbf{q}_{\text{gt}}) = 2 \cdot \text{acos}\left(\left|\left(\mathbf{q}_{\text{gt}}^{-1} \otimes \mathbf{q}_{\text{est}}\right)_w\right|\right).$$ (21)

Here, $(\cdot)_w$ denotes the scalar component of the quaternion, and $\otimes$ denotes quaternion multiplication. This angular error was reported in degrees.

These metrics provide the means to analyze the system's performance in different aspects in detail.

## 4.2 Dataset and Evaluation Scenarios

For the performance evaluation of the developed algorithm, the EuRoC MAV dataset was used. The EuRoC MAV dataset is widely recognized as a preferred dataset for the development and evaluation of VIO, VINS, and SLAM algorithms [76].

This dataset includes various scenarios that reflect the fundamental challenges encountered in real-world applications. Particularly, the Machine Hall (MH) sequences, recorded in an industrial environment, are noteworthy as they encompass real-world difficulties, such as complex structures, reflections from metallic surfaces, and movement in narrow spaces.

Table 1 presents characteristic metadata for the MH sequences, such as sequence distance, duration, average speed, and average angular velocity. Specifically, the MH04 and MH05 sequences involve rapid movements combined with low illumination conditions, positioning them among the most challenging scenarios for testing the accuracy of VIO algorithms.

**Fig. 4** Mutual correlation assessment of statistical metrics for challenging scenarios in the Graph-Based Odometry process

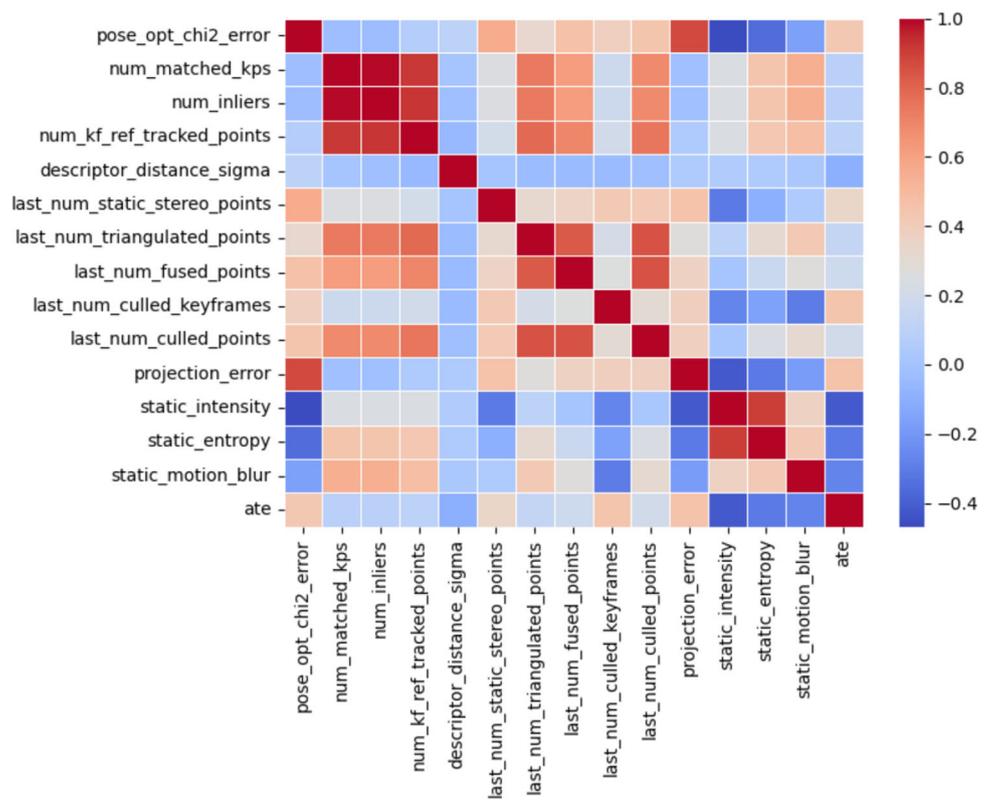

The adaptive Qf-ES-EKF/UKF algorithm proposed in our paper evaluates three critical metrics as prior information: intensity change, texture entropy, and motion blur. The selection of these metrics is not arbitrary; it is directly related to the fundamental challenges highlighted by the dataset. Within the scope of challenging scenarios, the correlation matrix for the detailed statistical analysis results of visual inferences from the MH04 and MH05 sequences is presented in Fig. 4.

The correlation matrix presented in Fig. 4 and detailed in Sect. 3.6 reveals noteworthy relationships, especially in challenging scenarios like MH04 and MH05. A significant positive correlation of 0.46 was observed between the ATE metric, reflecting position estimation accuracy, and *Pose Optimization* $\chi^2$ *Error*. A correlation of 0.42 was found between ATE and *Projection Error*. Due to the high correlation (0.88) between these two error metrics themselves, only one, namely *Pose Optimization* $\chi^2$ *Error*, was considered in the adaptive filter process.

Following the *Projection Error* metric, another important indicator showing the highest correlation with ATE (0.45) is the sudden decrease in the number of keyframes. The fundamental factors behind the preference for this metric in adaptive covariance updates are its strong correlation with ATE and its low correlation with the *Pose Optimization* $\chi^2$ *Error* metric. Thus, it is anticipated that it could benefit the system in certain rare scenarios.

Another important finding is that *static intensity* and normalized *static entropy* values exhibit negative correlations with ATE at levels of $-0.42$ and $-0.32$, respectively; this highlights the significant impact of illumination conditions on inference accuracy. Due to the calculation of the *static entropy* metric using the formula $1 - entropy_{\mathcal{N}}$, a high positive correlation of $+0.90$ was observed between this normalized *static entropy* and *static intensity*, contrary to the expected negative relationship. Finally, the negative correlation of $-0.27$ between motion blur and ATE is noteworthy. It is thought that this relationship can be explained by the observation of low ATE values despite high blur due to optical focus in scenarios where the system is stationary and the camera is positioned close to the scene. In this case, the low ATE is a result of stable measurements stemming from the system's stationarity rather than the blur, and this correlation has been assessed as a "false positive" effect.

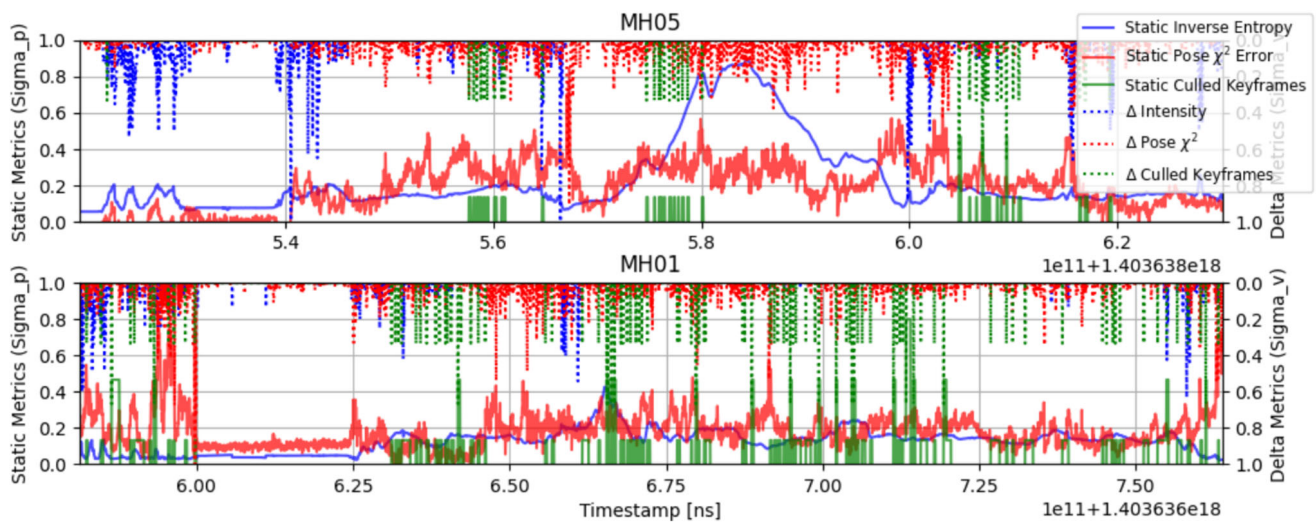

**Fig. 5** Graphical representation of metrics controlling visual position covariance ($\sigma_p$) (bottom to top) and visual velocity covariance ($\sigma_v$) (top to bottom) over time

Figure 5 visualizes key metrics for two distinct Machine Hall sequences: MH01 and MH05. The figure illustrates the static metrics associated with position covariance $\sigma_p$ (*static entropy*, *static motion blur*, *static Pose Optimization* $\chi^2$ *Error*), alongside the dynamic variables that affect velocity covariance $\sigma_v$ (*delta intensity*, *delta motion blur*, *delta Pose Optimization* $\chi^2$ *Error*, and *delta Culled Keyframes*). In the graph, static metrics are presented from bottom to top, while delta values are ranked from top to bottom. This visualization offers important insights into how the proposed algorithm evaluates visual measurements in dynamic scenes and adapts covariance updates.

Upon examining Fig. 5, the high metric values in the MH05 sequences indicate that the proposed adaptive mechanism effectively activates, especially in these challenging sequences. There is a fundamental reason for preferring the *static entropy* value over the *static intensity* metric in the adaptive determination of the $\sigma_p$ value. This criterion can produce more consistent thresholding levels across five different sequences compared to intensity values. Furthermore, it can keep the light level low in MH01, MH02, and MH03 sequences. Since *static entropy* shows a high correlation of approximately 0.90 with *static intensity*, only one was used to avoid redundancy in calculations.

### 4.3 Hyperparameter Optimization Strategy

The effectiveness of the proposed adaptive VIO system is highly dependent on the precise tuning of several critical hyperparameters. These include the confidence thresholds ($W_{thr}$, $D_{thr}$), the CASEF activation function parameter ($s$), and the dynamic confidence metric weights ($\alpha, \beta, \gamma, \zeta$). Finding the optimal set for these parameters is a challenging optimization problem due to their complex and non-linear interactions. To systematically address this problem and ensure consistent results across all experiments, we developed a two-stage hyperparameter optimization strategy, which is visualized in Fig. 6.

#### 4.3.1 Stage 1: Narrowing the Search Space via Stepwise Analysis

The primary goal of this initial stage is to establish an efficient starting point for the global optimization. The optimization efforts were focused on the weighting parameters $\alpha, \beta, \gamma, \zeta$, as they possessed the widest and most uncertain initial ranges. In contrast, other hyperparameters like $W_{thr}$ and $s$ already have more constrained and well-defined ranges based on theoretical considerations or practical observations. For instance, the initial value for the weighting threshold, $W_{thr}$, was specifically chosen at a level that would not trigger the adaptive mechanism in sequences where the visual odometry already performs well (e.g., "MH01", "MH02", "MH03"), but would activate it in more challenging sequences with poorer visual quality (e.g., "MH04", "MH05"). This ensures that the system avoids aggressive, unnecessary interventions in reliable conditions.

**Fig. 6** The two-stage hyperparameter optimization strategy. Stage 1 performs a coarse search to find good starting points and narrowed ranges, which are then used in Stage 2 to find a global optimum with a Genetic Algorithm (GA)

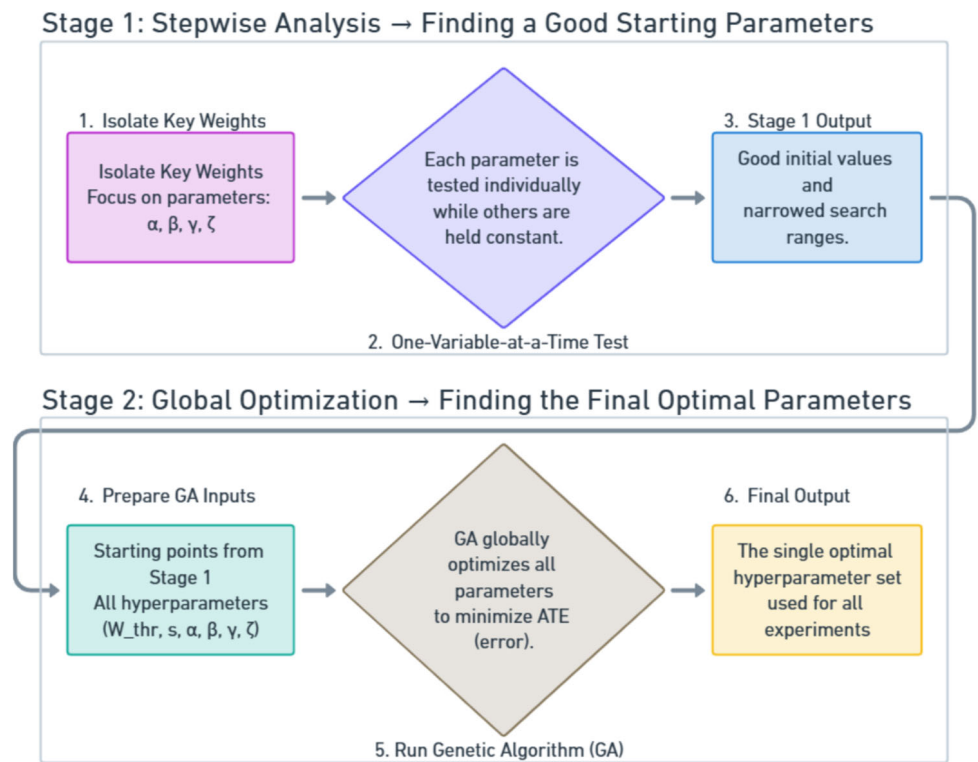

With these more predictable parameters held at reasonable initial values (e.g., $s = 1.0$, $W_{thr} = 0.2$), we employed a "One-Variable-at-a-Time" (OVAT) approach for the uncertain weight parameters. By analyzing the impact of each parameter on the system's ATE, we identified coarse ranges that yielded the best performance. The main advantage of this process is that it significantly narrows the search space for the more computationally expensive global optimization in the second stage. However, this stage cannot capture inter-parameter interactions and only provides a good initial guess.

### *Stage 2: Global Optimization with a Genetic Algorithm*

The narrowed search ranges obtained from the first stage form the foundation for the second and final stage: global optimization. In this stage, we used a Genetic Algorithm (GA)—a method well suited for such problems—to effectively model the complex, non-convex relationships between parameters and find a global optimum.

The GA was configured to simultaneously optimize the **entire set of hyperparameters** ($W_{thr}$, $s$, $\alpha$, $\beta$, $\gamma$, $\zeta$) by minimizing the ATE on related sequences from the EuRoC MAV dataset, which is considered one of the most challenging scenarios. The population-based search mechanism of the GA allows it to explore the parameter space broadly, reducing the risk of getting trapped in local minima. The process culminates in a **single, optimal hyperparameter set**, which was then used consistently across all experiments in our study. This systematic approach ensures the fairness of our comparisons and enhances the reproducibility and reliability of our findings.

To evaluate the effectiveness of the innovations introduced by the proposed filtering system, the ESKF algorithm, which is widely accepted in the literature and whose success has been extensively tested, was chosen as the reference method. ESKF is a filtering technique operating in the error-state space, frequently used especially in visual–inertial odometry systems. It is known for providing precise state estimation by modeling uncertainties in sensor measurements and for its suitability for real-time applications due to its low computational cost. Within this framework, the performance of the proposed method was quantitatively evaluated through comparative analyses with ESKF, and the contributions of the introduced innovations to system performance were demonstrated.

**Table 2** Comparative rotation accuracy metrics (Euler Angles and Quaternion RMSE) with overall averages

| Sequence | Method | Euler Angles | | | Quaternion RMSE (°) |
|---|---|---|---|---|---|
| | | Roll (°) | Pitch (°) | Yaw (°) | |
| MH05 | ESKF | **0.393533** | 0.461227 | 2.103856 | 2.164631 |
| | SUKF | 0.929114 | 0.506769 | 0.995790 | 0.945651 |
| | Qf-ES-EKF/UKF | 0.531587 | 0.467379 | 0.763492 | 0.837798 |
| | Adaptive Qf-ES-EKF/UKF | 0.431013 | **0.447647** | **0.444956** | **0.525513** |
| MH04 | ESKF | 0.842355 | 0.178753 | 0.887193 | 0.647382 |
| | SUKF | 0.895132 | 0.221625 | 0.690709 | 0.599839 |
| | Qf-ES-EKF/UKF | 0.708330 | **0.125699** | 0.689431 | 0.412316 |
| | Adaptive Qf-ES-EKF/UKF | **0.505267** | 0.128998 | **0.433787** | **0.313051** |
| MH03 | ESKF | 0.669203 | 0.155828 | 0.661883 | 0.360864 |
| | SUKF | 1.448759 | 0.365193 | 1.668660 | 0.755577 |
| | Qf-ES-EKF/UKF | 0.562030 | 0.122705 | 0.580660 | **0.250244** |
| | Adaptive Qf-ES-EKF/UKF | **0.533043** | **0.116171** | 0.572824 | 0.374603 |
| MH02 | ESKF | 1.648101 | 0.411688 | 1.872213 | 0.956780 |
| | SUKF | 1.563768 | **0.302952** | **1.668789** | **0.907337** |
| | Qf-ES-EKF/UKF | **1.475485** | 0.321505 | 1.778834 | 1.158065 |
| | Adaptive Qf-ES-EKF/UKF | 1.608951 | 0.319040 | 1.845856 | 1.126859 |
| MH01 | ESKF | 2.119460 | 0.888907 | 4.164797 | 2.976050 |
| | SUKF | 2.015549 | 0.492726 | 2.168348 | 1.036893 |
| | Qf-ES-EKF/UKF | 0.970420 | 0.688612 | 1.144351 | 1.056485 |
| | Adaptive Qf-ES-EKF/UKF | **0.893415** | **0.388368** | **1.068468** | **0.720242** |
| **Average** | ESKF | 1.134530 | 0.419281 | 1.937988 | 1.421141 |
| | SUKF | 1.370464 | 0.377853 | 1.438459 | 0.849059 |
| | Qf-ES-EKF/UKF | 0.849570 | 0.345180 | 0.991354 | 0.742982 |
| | Adaptive Qf-ES-EKF/UKF | **0.794338** | **0.280045** | **0.873178** | **0.612054** |

Table 2 includes a comparison of quaternion and Euler angles for ESKF, SUKF, and Adaptive Qf-ES-EKF/UKF methods. The comparison of Euler angles is given in degrees, not radians.

Table 3 presents a comparative overview of the ATE and axis-based RMSE performances of the ESKF and adaptive Qf-ES-EKF/UKF filters for five different sequences in the EuRoC MAV dataset. This analysis aims to evaluate the impact of the proposed adaptive mechanism on system performance by examining the position estimation accuracy of different filtering approaches.

Table 4 presents a comparative overview of the computation times for ESKF, SUKF, and Qf-ES-EKF/UKF algorithms operating with the basic state vector ($\mathbf{q}$, $\mathbf{p}$, $\mathbf{v}$, $\mathbf{b}_g$, $\mathbf{b}_a$) in experiments conducted on five different sequences from the EuRoC MAV dataset, without applying any parallelization techniques and with visual inferences directly integrated into the system. In real-time VIO systems, the balance between accuracy and computational cost is of great importance in algorithm selection. In this context, Table 4 provides a quantitative assessment of this balance, serving as a valuable reference for algorithm design and selection.

The quantitative performance of the proposed system across all five sequences is comprehensively detailed in Table 3. To provide a qualitative and more intuitive understanding of these results, Fig. 7 visualizes the estimated trajectories. To most effectively demonstrate the benefits of our adaptive mechanism, this visualization selectively focuses on the MH04 and MH05 sequences. These scenarios, characterized by poor illumination and rapid motion, represent the most challenging conditions in the dataset and are where the adaptive covariance adjustment has the most significant and discernible impact. In contrast, for the less demanding MH01, MH02, and MH03 sequences, both the proposed method and the baseline ESKF achieve high accuracy, producing trajectories that are visually almost indistinguishable from the ground truth. Consequently, their inclusion in the figure would offer little additional comparative insight while reducing visual clarity. As depicted for MH04 and MH05, the trajectory estimated by the adaptive Qf-ES-EKF/UKF exhibits a **markedly closer alignment** with the ground truth compared to the baseline ESKF, underscoring the system's **enhanced robustness** in adverse conditions.

**Table 3** Comparative position accuracy metrics: position RMSE and ATE

| Sequence | Method | Position RMSE (m) | | | ATE (m) |
|---|---|---|---|---|---|
| | | RMSE$_x$ | RMSE$_y$ | RMSE$_z$ | RMSE |
| MH05 | ESKF | 0.1843 | 0.4010 | **0.0993** | 0.4524 |
| | SUKF | 0.1830 | 0.4002 | 0.1009 | 0.4492 |
| | Qf-ES-EKF/UKF | 0.1817 | 0.3988 | 0.1002 | 0.4328 |
| | Adaptive Qf-ES-EKF/UKF | **0.1691** | **0.1129** | 0.1671 | **0.2632** |
| MH04 | ESKF | 0.2698 | 0.3493 | 0.3441 | 0.5597 |
| | SUKF | 0.2765 | 0.3390 | 0.3415 | 0.5558 |
| | Qf-ES-EKF/UKF | 0.2623 | 0.3489 | 0.3389 | 0.5286 |
| | Adaptive Qf-ES-EKF/UKF | **0.1228** | **0.1770** | **0.1288** | **0.2510** |
| MH03 | ESKF | **0.0400** | 0.0514 | 0.1296 | 0.1451 |
| | SUKF | 0.0564 | 0.0501 | 0.1085 | 0.1337 |
| | Qf-ES-EKF/UKF | 0.0412 | 0.0487 | 0.1274 | 0.1324 |
| | Adaptive Qf-ES-EKF/UKF | 0.0401 | **0.0470** | **0.1025** | **0.1197** |
| MH02 | ESKF | **0.0429** | 0.0547 | **0.0199** | **0.0724** |
| | SUKF | 0.0460 | 0.0530 | 0.0230 | 0.0754 |
| | Qf-ES-EKF/UKF | 0.0483 | **0.0516** | 0.0251 | 0.0786 |
| | Adaptive Qf-ES-EKF/UKF | 0.0487 | 0.0577 | 0.0686 | 0.1021 |
| MH01 | ESKF | 0.0413 | 0.0458 | 0.0818 | 0.0967 |
| | SUKF | 0.0408 | 0.0430 | 0.0785 | 0.0934 |
| | Qf-ES-EKF/UKF | **0.0404** | 0.0464 | 0.0753 | 0.0892 |
| | Adaptive Qf-ES-EKF/UKF | 0.0409 | **0.0431** | **0.0716** | **0.0854** |

**Table 4** Comparison of time consumption and real-time performance for the filtering algorithms across five sequences in the EuRoC MAV dataset. Real-Time Factor (RTF) is calculated as Sequence Duration/Processing Time. An RTF greater than 1.0 indicates real-time capability

| Sequence | Duration (s) | ESKF | | SUKF | | Qf-ES-EKF/UKF | |
|---|---|---|---|---|---|---|---|
| | | Time (s) | RTF | Time (s) | RTF | Time (s) | RTF |
| MH05 | 111 | 45.70 | **2.43** | 121.46 | *0.91* | 63.78 | **1.74** |
| MH04 | 99 | 40.80 | **2.43** | 108.09 | *0.92* | 56.64 | **1.75** |
| MH03 | 132 | 54.35 | **2.43** | 143.94 | *0.92* | 75.29 | **1.75** |
| MH02 | 150 | 61.46 | **2.44** | 164.19 | *0.91* | 85.67 | **1.75** |
| MH01 | 182 | 73.63 | **2.47** | 198.48 | *0.92* | 103.27 | **1.76** |
| **Average** | 134.80 | 55.19 | 2.44 | 147.23 | *0.92* | 76.93 | **1.75** |

The aggregate performance of our proposed method, summarized in Table 5, highlights its key advantages over baseline filters across three critical domains: rotation accuracy, positional robustness, and computational efficiency. The hybrid Qf-ES-EKF/UKF architecture delivers a substantial improvement in orientation estimation, reducing the average rotation error by 57% compared to the standard ESKF. This result validates our hypothesis that selectively applying the SUKF to the non-linear quaternion state effectively mitigates the linearization errors of traditional EKF-based approaches. Moreover, the adaptive mechanism demonstrates its most significant impact under challenging conditions where visual data quality is poor (MH04 and MH05). In these scenarios, our method

**Table 5** Summary of relative performance improvements of the proposed adaptive Qf-ES-EKF/UKF method compared to baseline filters. Positive values indicate a percentage improvement (lower error or less time), while negative values indicate a performance decrease

| Performance Metric | vs. ESKF | vs. SUKF | vs. Non-Ad. Qf-ES-EKF/UKF |
|---|---|---|---|
| Overall Rotation Accuracy[a] | **+57%** | +28% | +18% |
| Position Accuracy (Challenging)[b] | **+49%** | **+49%** | +47% |
| Computational Efficiency[a] | −39% | **+48%** | ≈ 0%[c] |

[a] Average improvement across all five EuRoC MH sequences
[b] Average improvement on challenging sequences (MH04 and MH05) only
[c] The adaptive mechanism adds negligible computational overhead to the core filter

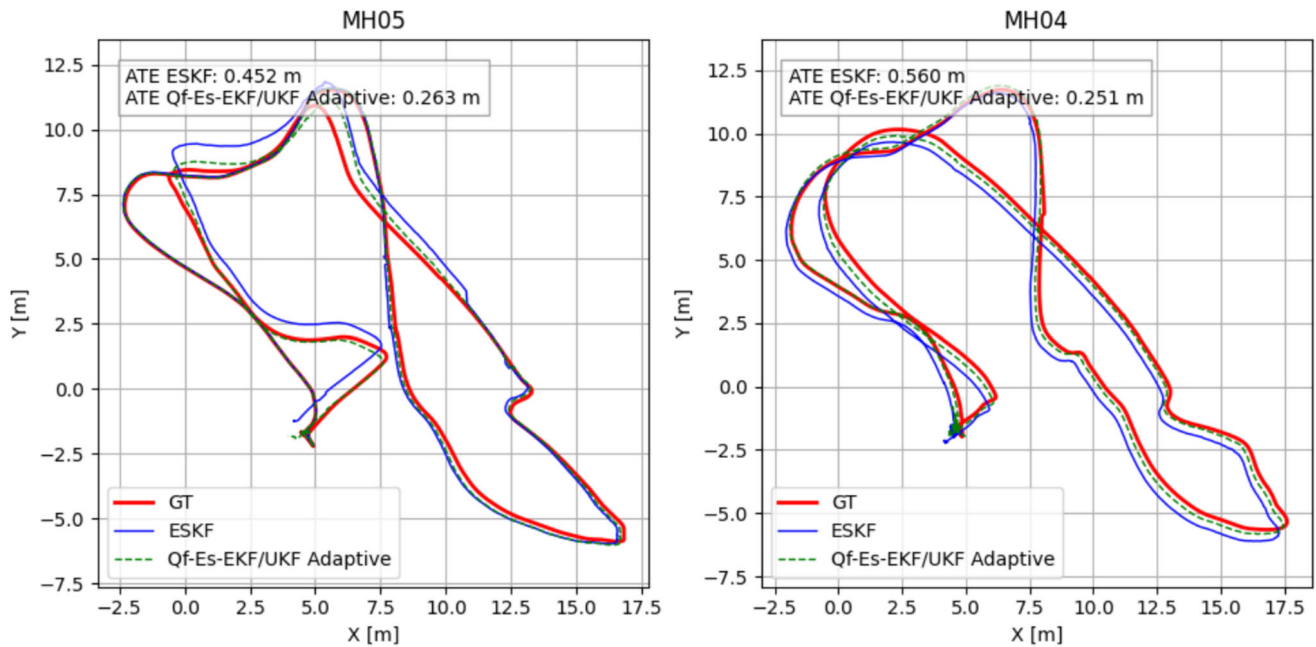

**Fig. 7** Trajectory comparison for the challenging MH04 and MH05 sequences. The visualization highlights the **improved accuracy** of the adaptive Qf-ES-EKF/UKF under adverse conditions. Less demanding sequences (MH01–MH03) are omitted as both methods produce trajectories nearly identical to the ground truth

reduces the average Absolute Trajectory Error (ATE) by a remarkable 49% relative to both the ESKF and the full SUKF. This enhanced positional robustness is directly attributable to the system's ability to dynamically down-weight unreliable visual measurements, a conclusion reinforced by the 47% improvement over its non-adaptive counterpart. Crucially, these accuracy and robustness gains are achieved without the computational penalty of higher-order filters. Our method operates approximately 48% faster than a full SUKF implementation, striking an effective balance that is essential for real-time VIO on resource-constrained platforms. With the adaptive layer adding negligible computational overhead, our approach stands as a practical and powerful solution for real-world navigation tasks.

## 5 Conclusions

This study introduced an innovative hybrid VIO approach for UAVs, featuring an adaptive covariance update mechanism designed for high performance in challenging environmental conditions. The proposed system's Quaternion-Focused Error-State EKF/UKF (Qf-ES-EKF/UKF) filter strikes a crucial balance by merging the superior modeling capability of SUKF for non-linear orientation with the computational efficiency of ESKF for other state variables.

Experimental results confirmed that our hybrid filter design overcomes the limitations of the standard ESKF in modeling the complex $SO(3)$ manifold, especially during highly dynamic movements. The superior orientation performance of our hybrid filter is a direct result of its design. For each IMU measurement, the filter first performs an ESKF-based propagation for the entire state, creating an initial estimate. It then refines the orientation component of this estimate with a UKF step, which more accurately handles the non-linear dynamics. This "two-stage" structure leverages the efficiency of ESKF while correcting for orientation errors with the accuracy of UKF, enabling it to produce more stable and precise results than even a standalone SUKF implementation.

When examining position estimation performance, a different dynamic emerged. For states with less non-linearity, such as position and velocity, ESKF and SUKF exhibited similar performance on the EuRoC dataset, where visual data are consistently available. Under these conditions, the modest advantage of SUKF in position estimation remained an indirect consequence of its superior orientation tracking. It is anticipated, however, that in real-world scenarios with prolonged visual outages, SUKF's ability to more accurately integrate inertial data would offer a clear advantage over ESKF in reducing accumulated position error (drift). In contrast, our adaptive mechanism addresses this challenge differently, delivering a radical improvement by directly evaluating and filtering the visual data—the primary source for position updates—based on real-time quality metrics. By mitigating the corrupting influence of low-quality or erroneous visual measurements at the source, a significant increase in position accuracy was achieved. The overall performance and relative improvements of the system are summarized in Table 5.

In our work, we also identified areas for future improvement. The approach for motion blur analysis, based on the variance of the *Laplacian operator*, was less effective than anticipated on the global shutter imagery from the EuRoC MAV dataset. This suggests that the metric may be more impactful in datasets featuring rolling shutter cameras or more dynamic motion profiles. Future work could explore more robust methods for blur detection, such as those based on optical flow or deep learning, as a viable research direction.

The balance between computational efficiency and estimation accuracy is critical for VIO systems. Our time consumption analysis shows that the proposed hybrid approach successfully strikes this balance. While the hybrid filter is significantly faster than a full SUKF implementation, its computational overhead compared to the standard ESKF is reasonable, especially considering the substantial accuracy gains. This demonstrates that our approach holds strong potential for real-time applications.

In conclusion, the adaptive and hybrid VIO system presented in this study offers significant contributions to the literature. It provides a computationally efficient architecture that enhances both orientation and position accuracy by intelligently adapting to varying sensor reliability. This approach has the potential to significantly improve the navigation performance of UAVs, which is a foundational requirement for advanced autonomous tasks. Although the parameters of the adaptive mechanism were optimized for specific datasets, the fundamental framework is flexible and robust, providing a solid foundation for future advancements in VIO systems.

# 6 Future Work

Future work is planned to focus on the following topics:

- Adaptive particle count optimization for particle filter: It is aimed to develop an algorithm that dynamically adjusts the number of particles based on image quality. This could provide an optimal balance between computational efficiency and estimation accuracy.
- Extended confidence metric for multi-sensor fusion: The current approach is planned to be extended to include additional sensors, such as LiDAR, GPS, and 5 G. This will enable the creation of a more comprehensive and robust sensor fusion framework.

- Deep learning-based confidence estimation: The use of deep learning models for assessing image quality and information content is targeted. This use could make the confidence metric more sophisticated and context-aware. Furthermore, it could eliminate the process of finding correct parameters.
- Real-time application optimization: Work is planned on algorithm optimizations and parallel processing techniques to increase the computational efficiency of the proposed approach and improve its usability in real-time applications.

These future studies aim to expand the scope and enhance the applicability of the proposed image-based adaptive sensor fusion approach.

**Acknowledgements** This research did not receive any specific grant from funding agencies in the public, commercial, or not-for-profit sectors. The author(s) declare that they have no known competing financial interests or personal relationships that could have appeared to influence the work reported in this paper.

**Author Contributions** UA conducted all the experiments and wrote the entire article. EN supervised the experiments and the article.

**Funding** This research did not receive external funding.

**Data Availability** The data used in this study are publicly available as part of the EuRoC MAV (Micro-aerial Vehicle) dataset. No confidential or proprietary data were used in this research. The EuRoC MAV dataset can be accessed at http://robotics.ethz.ch. Any additional materials or code used for analysis are available from the corresponding author upon reasonable request.

## Declarations

**Conflict of Interest** The authors declare that they have no relevant financial or non-financial interests to disclose.

**Ethical Approval** The authors are committed to adhering to the Committee on Publication Ethics (COPE) guidelines.

## References

1. Li, X., Gao, F., Zhao, C.: Smart path planning for UAV navigation using deep reinforcement learning in gis environment. Int. J. Comput. Intell. Syst. **14**(1), 47–59 (2021). https://doi.org/10.1007/s44196-021-00031-y
2. Choi, H.-B., Lim, K.-W., Ko, Y.-B.: LUVI: lightweight UWB-VIO based relative positioning for AR-IoT applications. Ad Hoc Netw. **145**, 103132 (2023). https://doi.org/10.1016/j.adhoc.2023.103132
3. Li, J., Zhou, X., Yang, B., Zhang, G., Wang, X., Bao, H.: Rlp-vio: Robust and lightweight plane-based visual-inertial odometry for augmented reality. Comput. Anim. Virtual Worlds **34**(2), 21106 (2023). https://doi.org/10.1002/cav.21106
4. Yu, Z., Zhu, L., Lu, G.: VINS-Motion: Tightly-coupled fusion of VINS and motion constraint. In: 2021 IEEE International Conference on Robotics and Automation (ICRA), pp. 7672–7678 (2021). https://doi.org/10.1109/ICRA48506.2021.9562103
5. Mostafa, M., Zahran, S., Moussa, A., El-Sheimy, N., Sesay, A.: Radar and visual odometry integrated system aided navigation for UAVS in GNSS denied environment. Sensors **18**(9), 2776 (2018). https://doi.org/10.3390/s18092776
6. Liu, Y., Zhang, Y., Liang, P., Fu, Y.: Yolov8n-based small object detection method for UAV images. Int. J. Comput. Intell. Syst. **17**(1), 84 (2024). https://doi.org/10.1007/s44196-024-00632-3
7. Sarlin, P.-E., DeTone, D., Malisiewicz, T., Rabinovich, A.: Superglue: Learning feature matching with graph neural networks. In: 2020 IEEE/CVF Conference on Computer Vision and Pattern Recognition (CVPR), pp. 4937–4946 (2020). https://doi.org/10.1109/CVPR42600.2020.00499

8. Lindenberger, P., Sarlin, P.-E., Pollefeys, M.: Lightglue: Local feature matching at light speed. In: 2023 IEEE/CVF International Conference on Computer Vision (ICCV), pp. 17581–17592 (2023). https://doi.org/10.1109/ICCV51070.2023.01616

9. Sun, J., Shen, Z., Wang, Y., Bao, H., Zhou, X.: LoFTR: detector-free local feature matching with transformers. 2021 IEEE/CVF Conference on Computer Vision and Pattern Recognition (CVPR), 8918–8927 (2021) https://doi.org/10.1109/cvpr46437.2021.00881

10. Jiang, W., Trulls, E., Hosang, J., Tagliasacchi, A., Yi, K.M.: Correspondence transformer for matching across images. In: 2021 IEEE/CVF International Conference on Computer Vision (ICCV), pp. 6187–6197 (2021). https://doi.org/10.1109/ICCV48922.2021.00615

11. Wang, Q., Zhang, J., Yang, K., Peng, K., Stiefelhagen, R.: MatchFormer: Interleaving attention in transformers for feature matching. In: Proc. Asian Conf. Comput. Vis. (ACCV), pp. 256–273. Springer, Cham (2022). https://doi.org/10.1007/978-3-031-26313-2_16

12. Li, L.-Q., Ji, H.-B., Luo, J.-H.: The iterated extended Kalman particle filter. In: Proc. IEEE Int. Symp. Commun. Inf. Technol. (ISCIT), vol. 2, pp. 1213–1216 (2005). https://doi.org/10.1109/ISCIT.2005.1567087

13. Wan, E.A., van der Merwe, R.: The unscented Kalman filter for nonlinear estimation. In: Proc. IEEE 2000 Adapt. Syst. Signal Process. Commun. Control Symp., pp. 153–158 (2000). https://doi.org/10.1109/ASSPCC.2000.882463

14. Bescos, B., Campos, C., Tardós, J.D., Neira, J.: Dynaslam ii: tightly-coupled multi-object tracking and slam. IEEE Robot. Autom. Lett. **6**(3), 5191–5198 (2021). https://doi.org/10.1109/LRA.2021.3068640

15. Huang, J., Wen, S., Liang, W., Guan, W.: VWR-SLAM: tightly coupled SLAM system based on visible light positioning landmark, wheel odometer, and RGB-D camera. IEEE Trans. Instrum. Meas. **72**, 1–12 (2023). https://doi.org/10.1109/TIM.2022.3231332

16. Liang, Q., Liu, M.: A tightly coupled VLC-inertial localization system by EKF. IEEE Robot. Autom. Lett. **5**(2), 3129–3136 (2020). https://doi.org/10.1109/LRA.2020.2975730

17. Leutenegger, S., Lynen, S., Bosse, M., Siegwart, R., Furgale, P.: Keyframe-based visual inertial odometry using nonlinear optimization. Int. J. Robot. Res. **34**(3), 314–334 (2015). https://doi.org/10.1177/0278364914554813

18. Forster, C., Carlone, L., Dellaert, F., Scaramuzza, D.: On-manifold preintegration for real-time visual-inertial odometry. IEEE Trans. Rob. **33**(1), 1–21 (2017). https://doi.org/10.1109/TRO.2016.2597321

19. Glukhov, O.V., Akinfiev, I.A., Razorvin, A.D., Chugunov, A.A., Gutarev, D.A., Serov, S.A.: Loosely coupled UWB/stereo camera integration for mobile robots indoor navigation. In: 2023 5th International Youth Conference on Radio Electronics, Electrical and Power Engineering (REEPE), vol. 5, pp. 1–7 (2023). https://doi.org/10.1109/REEPE57272.2023.10086807

20. Cen, R., Liu, S., Xue, F.: 3d mapping based imu loosely coupled model for autonomous robot. In: 2021 IEEE International Conference on Intelligence and Safety for Robotics (ISR), pp. 196–199 (2021). https://doi.org/10.1109/ISR50024.2021.9419513

21. Lee, S.H., Civera, J.: Loosely-coupled semi-direct monocular slam. IEEE Robot. Autom. Lett. **4**(2), 399–406 (2019). https://doi.org/10.1109/LRA.2018.2889156

22. Sirtkaya, S., Seymen, B., Alatan, A.A.: Loosely coupled kalman filtering for fusion of visual odometry and inertial navigation. In: Proceedings of the 16th International Conference on Information Fusion, pp. 219–226. IEEE, Istanbul, Turkey (2013)

23. Gopaul, N.S., Wang, J., Hu, B.: Loosely coupled visual odometry aided inertial navigation system using discrete extended Kalman filter with pairwise time correlated measurements. In: 2017 Forum on Cooperative Positioning and Service (CPGPS), pp. 283–288 (2017). https://doi.org/10.1109/cpgps.2017.8075140

24. Konolige, K., Agrawal, M., Solà, J.: Large-scale visual odometry for rough terrain. In: Kaneko, M., Nakamura, Y. (Eds.) Robotics Research, pp. 201–212. Springer, Berlin, Heidelberg (2011). https://doi.org/10.1007/978-3-642-14743-2_18

25. Shen, S., Mulgaonkar, Y., Michael, N., Kumar, V.: Multi-sensor fusion for robust autonomous flight in indoor and outdoor environments with a rotorcraft mav. In: 2014 IEEE International Conference on Robotics and Automation (ICRA), pp. 4974–4981 (2014). https://doi.org/10.1109/ICRA.2014.6907588

26. Scaramuzza, D., Achtelik, M.C., Doitsidis, L., Friedrich, F., Kosmatopoulos, E., Martinelli, A., Achtelik, M.W., Chli, M., Chatzichristofis, S., Kneip, L., et al.: Vision-controlled micro flying robots: from system design to autonomous navigation and mapping in GPS-denied environments. IEEE Robot. Autom. Magaz. **21**(3), 26–40 (2014). https://doi.org/10.1109/MRA.2014.2322295

27. Kabiri, M., Cimarelli, C., Bavle, H., Sanchez-Lopez, J.L., Voos, H.: Graph-based vs. error state Kalman filter-based fusion of 5g and inertial data for MAV indoor pose estimation. J. Intell. Robot. Syst. **110**(2), 87 (2024). https://doi.org/10.1007/s10846-024-02111-5

28. Zhou, T., Yang, M., Jiang, K., Wong, H.T., Yang, D.: Mmw radar-based technologies in autonomous driving: a review. Sensors **20**, 7283 (2020). https://doi.org/10.3390/s20247283

29. Xu, X., Zhang, L., Yang, J., Cao, C., Wang, W., Ran, Y., Tan, Z., Luo, M.: A review of multi-sensor fusion SLAM systems based on 3d LIDAR. Remote Sens. (2022). https://doi.org/10.3390/rs14122835

30. Ye, B., Zhang, H., Shi, Y.: Qoblo: Quality-of-service aware opposition-based optimization algorithm for drone-based stations in 6g networks. Int. J. Comput. Intell. Syst. **17**(1), 78 (2024). https://doi.org/10.1007/s44196-024-00628-z

31. Mourikis, A.I., Roumeliotis, S.I.: A multi-state constraint Kalman filter for vision-aided inertial navigation. In: Proceedings 2007 IEEE International Conference on Robotics and Automation, pp. 3565–3572 (2007). https://doi.org/10.1109/ROBOT.2007.364024

32. Weiss, S., Achtelik, M.W., Lynen, S., Chli, M., Siegwart, R.: Real-time onboard visual-inertial state estimation and self-calibration of mavs in unknown environments. In: 2012 IEEE International Conference on Robotics and Automation, pp. 957–964 (2012). https://doi.org/10.1109/ICRA.2012.6225147

33. Fan, Z., Hao, Y., Zhi, T., Guo, Q., Du, Z.: Hardware acceleration for slam in mobile systems. J. Comput. Sci. Technol. **38**, 1300–1322 (2023). https://doi.org/10.1007/s11390-021-1523-5

34. Jane, H.: Comparison of ekf, ekf2 and ukf in a loosely coupled ins/GPS integration. Master's thesis, Middle East Technical University (Turkey) (2018). https://open.metu.edu.tr/bitstream/handle/11511/27047/index.pdf

35. Kurt-Yavuz, Z., Yavuz, S.: A comparison of EKF, UKF, FastSLAM2.0, and UKF-based FastSLAM algorithms. In: 2012 IEEE 16th International Conference on Intelligent Engineering Systems (INES), pp. 37–43 (2012). https://doi.org/10.1109/INES.2012.6249866

36. Wu, Y., Hu, D., Wu, M., Hu, X.: Unscented Kalman filtering for additive noise case: augmented vs. non-augmented. In: Proceedings of the 2005, American Control Conference, 2005., pp. 4051–4055 (2005). https://doi.org/10.1109/ACC.2005.1470611

37. Urrea, C., Agramonte, R.: Kalman filter: historical overview and review of its use in robotics 60 years after its creation. J. Sensors (2021). https://doi.org/10.1155/2021/9674015

38. Zhu, J., Zheng, N., Yuan, Z., Zhang, Q., Zhang, X., He, Y.: A slam algorithm based on the central difference Kalman filter. In: 2009 IEEE Intelligent Vehicles Symposium, pp. 123–128 (2009). https://doi.org/10.1109/IVS.2009.5164264

39. Zhang, X.-C.: A novel cubature Kalman filter for nonlinear state estimation. In: 52nd IEEE Conference on Decision and Control, pp. 7797–7802 (2013). https://doi.org/10.1109/CDC.2013.6761127

40. Luo, X., Moroz, I.M., Hoteit, I.: Scaled unscented transform gaussian sum filter: theory and application. Physica D **239**(10), 684–701 (2010). https://doi.org/10.1016/j.physd.2010.01.022

41. Merwe, R., Wan, E.A.: The square-root unscented Kalman filter for state and parameter-estimation. In: 2001 IEEE International Conference on Acoustics, Speech, and Signal Processing. Proceedings (Cat. No.01CH37221), vol. 6, pp. 3461–3464 (2001). https://doi.org/10.1109/ICASSP.2001.940586

42. Ito, K., Xiong, K.: Gaussian filters for nonlinear filtering problems. IEEE Trans. Autom. Control **45**(5), 910–927 (2000). https://doi.org/10.1109/9.855552

43. Hou, B., He, Z., Li, D., Zhou, H., Wang, J.: Maximum correntropy unscented Kalman filter for ballistic missile navigation system based on sins/CNS deeply integrated mode. Sensors (2018). https://doi.org/10.3390/s18061724

44. Zhao, J., Mili, L.: A theoretical framework of robust h-infinity unscented Kalman filter and its application to power system dynamic state estimation. IEEE Trans. Signal Process. **67**(10), 2734–2746 (2019). https://doi.org/10.1109/TSP.2019.2908910

45. Taghvaei, A., Mehta, P.G., Georgiou, T.T.: Optimality vs stability trade-off in ensemble Kalman filters. IFAC-PapersOnLine **55**(30), 335–340 (2022). https://doi.org/10.1016/j.ifacol.2022.11.075

46. Brossard, M., Bonnabel, S., Barrau, A.: Unscented Kalman filter on lie groups for visual inertial odometry. In: 2018 IEEE/RSJ International Conference on Intelligent Robots and Systems (IROS), pp. 649–655 (2018). https://doi.org/10.1109/IROS.2018.8593627

47. Chen, W., Yang, Y., Liu, S., Sun, W.: Hybrid algorithm of ant colony and sparrow search for UAV path planning problem. Int. J. Comput. Intell. Syst. **17**(1), 104 (2024). https://doi.org/10.1007/s44196-024-00652-z

48. Xu, X., Wang, X., Liu, Y., Li, C.: 3d path planning for UCAV based on improved plasma algorithm. Int. J. Comput. Intell. Syst. **16**(1), 112 (2023). https://doi.org/10.1007/s44196-023-00284-9

49. Li, H., Han, J., Liu, R., Ren, Q.: Time-constrained three-dimensional multi-UAV task planning based on improved ant colony optimization algorithm. Int. J. Comput. Intell. Syst. **14**(2), 181–193 (2021). https://doi.org/10.2991/ijcis.d.201021.001

50. Wang, X., Zhao, Y., Zhang, Y., Zhou, B.: Task allocation strategy for multiple heterogeneous UAVS based on orchard picking algorithm. Int. J. Comput. Intell. Syst. **14**(2), 240–250 (2021). https://doi.org/10.2991/ijcis.d.210423.003

51. Mur-Artal, R., Tardós, J.D.: ORB-SLAM2: an open-source SLAM system for monocular, stereo and RGB-D cameras. IEEE Trans. Rob. **33**(5), 1255–1262 (2017). https://doi.org/10.1109/TRO.2017.2705103

52. Li, L., Zhang, Z., Wang, Z., Xue, X.: Loop closure detection with lightweight convolutional neural networks in visual slam for indoor environments. Int. J. Comput. Intell. Syst. **16**(1), 105 (2023). https://doi.org/10.1007/s44196-023-00223-8

53. DeTone, D., Malisiewicz, T., Rabinovich, A.: Superpoint: Self-supervised interest point detection and description. In: Proceedings of the IEEE Conference on Computer Vision and Pattern Recognition (CVPR) Workshops (2018). https://doi.org/10.48550/arXiv.1712.07629

54. Dusmanu, M., Rocco, I., Pajdla, T., Pollefeys, M., Sivic, J., Torii, A., Sattler, T.: D2-net: a trainable CNN for joint description and detection of local features. In: 2019 IEEE/CVF Conference on Computer Vision and Pattern Recognition (CVPR), pp. 8092–8101 (2019). https://doi.org/10.1109/CVPR.2019.00828

55. Revaud, J., Weinzaepfel, P., Souza, C., Humenberger, M.: R2D2: Repeatable and reliable detector and descriptor. In: Wallach, H., Larochelle, H., Beygelzimer, A., d'Alché-Buc, F., Fox, E., Garnett, R. (eds.) Advances in Neural Information Processing Systems 32 (NeurIPS 2019), pp. 10134–10143. Curran Associates Inc, Red Hook, NY, USA (2019)

56. Zhao, X., Wu, X., Miao, J., Chen, W., Chen, P.C.Y., Li, Z.: ALIKE: Accurate and lightweight keypoint detection and descriptor extraction. IEEE Trans. Multimedia 25, 3101–3112 (2023). https://doi.org/10.1109/TMM.2022.3155927

57. Qin, T., Li, P., Shen, S.: Vins-mono: a robust and versatile monocular visual-inertial state estimator. IEEE Trans. Rob. 34(4), 1004–1020 (2018). https://doi.org/10.1109/TRO.2018.2853729

58. Leutenegger, S., Lynen, F., Bosse, M., Siegwart, R., Furgale, P.: Okvis: Open keyframe-based visual-inertial slam. In: Proceedings of the IEEE/RSJ International Conference on Intelligent Robots and Systems, pp. 3072–3079 (2015). https://doi.org/10.1109/IROS.2015.7360215

59. Rosinol, A., Abate, M., Chang, Y., Carlone, L.: Kimera: an open-source library for real-time metric-semantic localization and mapping. In: 2020 IEEE International Conference on Robotics and Automation (ICRA), pp. 1689–1696 (2020). https://doi.org/10.1109/ICRA40945.2020.9196885

60. Kaess, M., Johannsson, H., Roberts, R., Ila, V., Leonard, J.J., Dellaert, F.: iSAM2: incremental smoothing and mapping using the bayes tree. Int. J. Robot. Res. 31(2), 216–235 (2012). https://doi.org/10.1177/0278364911430419

61. Freda, L.: pySLAM: An open-source, modular, and extensible framework for SLAM (2025). https://arxiv.org/abs/2502.11955

62. Karpenko, A., Jacobs, D., Baek, J., Levoy, M.: Digital video stabilization and rolling shutter correction using gyroscopes. Technical Report CSTR 2011-03, Stanford University (2011). https://graphics.stanford.edu/papers/stabilization/karpenko_gyro.pdf

63. Hwangbo, M., Kim, J.-S., Kanade, T.: Gyro-aided feature tracking for a moving camera: fusion, auto-calibration and GPU implementation. Int. J. Robot. Res. 30(14), 1755–1774 (2011). https://doi.org/10.1177/0278364911416391

64. Chandraker, M., Agrawal, A., Kriegman, D., Belongie, S.: Motion blur in rolling shutter cameras. In: Proceedings of the IEEE Conference on Computer Vision and Pattern Recognition Workshops, pp. 780–787 (2014). https://doi.org/10.1109/CVPRW.2014.119

65. Guo, C., Kottas, D., DuToit, R., Ahmed, A., Li, R., Roumeliotis, S.: Hybrid visual-inertial tracking in a smartphone. J. Vis. Commun. Image Represent. 53, 205–222 (2018). https://doi.org/10.1016/j.jvcir.2018.03.013

66. Liu, Y., Ye, Z., Wei, Y., Lai, Z., Liu, W.: Robust image feature matching via progressive sparse spatial consensus. In: 2019 IEEE International Conference on Multimedia and Expo (ICME), pp. 1672–1677 (2019). https://doi.org/10.1109/ICME.2019.00288

67. Huang, W., Liu, H.: A robust pixel-aware gyro-aided KLT feature tracker for large camera motions. IEEE Trans. Instrum. Meas. 71, 1–14 (2022). https://doi.org/10.1109/TIM.2021.3129493

68. Kim, J., Kim, A.: Light condition invariant visual slam via entropy based image fusion. In: 2017 14th International Conference on Ubiquitous Robots and Ambient Intelligence (URAI), pp. 529–533 (2017). https://doi.org/10.1109/URAI.2017.7992661

69. Han, B., Lin, Y., Dong, Y., Wang, H., Zhang, T., Liang, C.: Camera attributes control for visual odometry with motion blur awareness. IEEE/ASME Trans. Mechatron. 28(4), 2225–2235 (2023). https://doi.org/10.1109/TMECH.2023.3234316

70. Zhao, W., Liang, R., Tan, X., Ma, Y.: Autonomous takeoff and landing of a drone using deep reinforcement learning and aruco marker. Int. J. Comput. Intell. Syst. 13(1), 17–28 (2020). https://doi.org/10.2991/ijcis.d.200615.002

71. Sola, J.: Quaternion kinematics for the error-state Kalman filter. Technical report, arXiv (2017). https://arxiv.org/abs/1711.02508

72. El-Diasty, M., Pagiatakis, S.: A rigorous temperature-dependent stochastic modelling and testing for mems-based inertial sensor errors. Sensors 9(11), 8473–8489 (2009). https://doi.org/10.3390/s91108473

73. Wu, Y., Chen, S.: An enhanced stochastic error modeling using multi-gauss Markov processes for GNSS/ins integration system. J. Eng. Appl. Sci. 71(1), 186 (2024). https://doi.org/10.1186/s44147-024-00520-9

74. Van Loan, C.: Computing integrals involving the matrix exponential. IEEE Trans. Autom. Control 23(3), 395–404 (1978). https://doi.org/10.1109/TAC.1978.1101743

75. Bar-Shalom, Y., Li, X.R., Kirubarajan, T.: State estimation for nonlinear dynamic systems. In: Estimation with Applications to Tracking and Navigation: Principles and Techniques, pp. 371–420. John Wiley & Sons, New York, NY, USA (2001)

76. Burri, M., Nikolic, J., Gohl, P., Schneider, T., Rehder, J., Omari, S., Achtelik, M., Siegwart, R.: The Euroc micro aerial vehicle datasets. Int. J. Robot. Res. 35, 1157–1163 (2016). https://doi.org/10.1177/0278364915620033

77. Geiger, A., Lenz, P., Stiller, C., Urtasun, R.: Vision meets robotics: The KITTI dataset. Int. J. Robot. Res. 32(11), 1231–1237 (2013). https://doi.org/10.1177/0278364913491297

78. Schubert, D., Goll, T., Demmel, N., Usenko, V., Stückler, J., Cremers, D.: The tum vi benchmark for evaluating visual-inertial odometry. In: 2018 IEEE/RSJ International Conference on Intelligent Robots and Systems (IROS), pp. 1680–1687 (2018). https://doi.org/10.1109/IROS.2018.8593419

79. Otsu, N.: A threshold selection method from gray-level histograms. IEEE Trans. Syst. Man Cybern. 9(1), 62–66 (1979). https://doi.org/10.1109/TSMC.1979.4310076

80. Kapur, J.N., Sahoo, P.K., Wong, A.K.C.: A new method for gray-level picture thresholding using the entropy of the histogram. Comput. Vis. Graph. Image Process. **29**(3), 273–285 (1985). https://doi.org/10.1016/0734-189X(85)90125-2

81. Peli, E.: Contrast in complex images. J. Opt. Soc. Am. A **7**(10), 2032–2040 (1990). https://doi.org/10.1364/JOSAA.7.002032

82. Remeš, V., Haindl, M.: Region of interest contrast measures. Kybernetika **54**(5), 978–990 (2018). https://doi.org/10.14736/kyb-2018-5-0978

83. Field, D.J.: Relations between the statistics of natural images and the response properties of cortical cells. J. Opt. Soc. Am. A **4**(12), 2379–2394 (1987). https://doi.org/10.1364/JOSAA.4.002379

84. Mannos, J., Sakrison, D.: The effects of a visual fidelity criterion of the encoding of images. IEEE Trans. Inf. Theory **20**(4), 525–536 (1974). https://doi.org/10.1109/TIT.1974.1055250

85. Ren, Y., Sun, L., Wu, G., Huang, W.: Dibr-synthesized image quality assessment based on local entropy analysis. In: 2017 International Conference on the Frontiers and Advances in Data Science (FADS), pp. 86–90 (2017). https://doi.org/10.1109/FADS.2017.8253200

86. Deng, H., Sun, X., Liu, M., Ye, C., Zhou, X.: Infrared small-target detection using multiscale gray difference weighted image entropy. IEEE Trans. Aerosp. Electron. Syst. **52**(1), 60–72 (2016). https://doi.org/10.1109/TAES.2015.140878

87. Chen, X., Zhang, Q., Lin, M., Yang, G., He, C.: No-reference color image quality assessment: from entropy to perceptual quality. EURASIP J. Image Video Process. (2019). https://doi.org/10.1186/s13640-019-0479-7

88. Grimaldi, D., Kurylyak, Y., Lamonaca, F.: Detection and parameters estimation of locally motion blurred objects. In: Proceedings of the 6th IEEE International Conference on Intelligent Data Acquisition and Advanced Computing Systems, vol. 1, pp. 483–487 (2011). https://doi.org/10.1109/IDAACS.2011.6072800

89. Tian, L., Qiu, K., Zhao, Y., Wang, P.: Edge detection of motion-blurred images aided by inertial sensors. Sensors (2023). https://doi.org/10.3390/s23167187

90. Chen, Y., Huang, J., Wang, J., Xie, X.: Edge Prior Augmented Networks for Motion Deblurring on Naturally Blurry Images (2021). https://arxiv.org/abs/2109.08915

91. Guo, Y.-Q., Gu, M., Xu, Z.-D.: Research on the improvement of semi-global matching algorithm for binocular vision based on lunar surface environment. Sensors (2023). https://doi.org/10.3390/s23156901

92. Kuemmerle, R., Grisetti, G., Strasdat, H., Konolige, K., Burgard, W.: g2o: A general framework for graph optimization. In: Proceedings of the IEEE International Conference on Robotics and Automation (ICRA), pp. 3607–3613 (2011). https://doi.org/10.1109/ICRA.2011.5979949

93. Grupp, M.: evo: Python package for the evaluation of odometry and SLAM. https://github.com/MichaelGrupp/evo (2017)

## Authors and Affiliations

**Ufuk Asil[1]** · **Efendi Nasibov[1]**

✉ Ufuk Asil
u.asil@ogr.deu.edu.tr

Efendi Nasibov
efendi.nasibov@deu.edu.tr

1    Department of Computer Science, Dokuz Eylul University, Buca, 35160 İzmir, Turkey